\documentclass{article} 
\usepackage{iclr2027_conference,times}

\usepackage{amsmath,amsfonts,bm}

\def\eqref#1{equation~\ref{#1}}

\def\1{\bm{1}}

\def\vone{{\bm{1}}}

\def\vb{{\bm{b}}}

\def\mA{{\bm{A}}}
\def\mB{{\bm{B}}}

\def\mL{{\bm{L}}}
\def\mM{{\bm{M}}}

\def\mR{{\bm{R}}}
\def\mS{{\bm{S}}}

\def\mU{{\bm{U}}}
\def\mV{{\bm{V}}}
\def\mW{{\bm{W}}}
\def\mX{{\bm{X}}}
\def\mY{{\bm{Y}}}

\def\mSigma{{\bm{\Sigma}}}

\DeclareMathAlphabet{\mathsfit}{\encodingdefault}{\sfdefault}{m}{sl}
\SetMathAlphabet{\mathsfit}{bold}{\encodingdefault}{\sfdefault}{bx}{n}
\newcommand{\tens}[1]{\bm{\mathsfit{#1}}}

\def\tH{{\tens{H}}}

\def\tV{{\tens{V}}}
\def\tW{{\tens{W}}}
\def\tX{{\tens{X}}}
\def\tY{{\tens{Y}}}
\def\tZ{{\tens{Z}}}

\usepackage{graphicx,booktabs,multirow}
\usepackage{placeins}
\usepackage[table]{xcolor}
\usepackage{arydshln}
\usepackage{subcaption}
\usepackage{amssymb}
\usepackage{capt-of}
\definecolor{linkcolor}{HTML}{ED1C24}
\definecolor{mydarkgreen}{rgb}{0.02,0.6,0.02}
\usepackage[colorlinks,citecolor=mydarkgreen,linkcolor=linkcolor]{hyperref}
\hypersetup{
  pdftitle={LightVAE: Towards Compact and Efficient Video Autoencoders},
  pdfauthor={Shiqiao Gu, Yang Yong, Fengjun Zhong, Ruihao Gong, Chengtao Lv, Xiangyu Fan, Lei Yang, Xianglong Liu}
}
\usepackage{url}

\title{LightVAE: Towards Compact and Efficient Video Autoencoders}

\author{%
  \textbf{Shiqiao Gu}$^{1}$ \quad
  \textbf{Yang Yong}$^{1}$ \quad
  \textbf{Fengjun Zhong}$^{2}$ \quad
  \textbf{Ruihao Gong}$^{2}$\thanks{Corresponding Author.} \\
  \textbf{Chengtao Lv}$^{3}$ \quad
  \textbf{Xiangyu Fan}$^{1}$ \quad
  \textbf{Lei Yang}$^{1}$ \quad
  \textbf{Xianglong Liu}$^{2}$ \\[3pt]
  {\normalfont\small $^{1}$SenseTime \quad
  $^{2}$Beihang University \quad
  $^{3}$Nanyang Technological University} \\[2pt]
  {\normalfont\small\texttt{\{gushiqiao,yongyang,fanxiangyu,yanglei\}@sensetime.com}} \\
  {\normalfont\small\texttt{\{fengjunzhong,gongruihao,xlliu\}@buaa.edu.cn}} \\
  {\normalfont\small\texttt{chengtao001@e.ntu.edu.sg}}
}

\iclrfinalcopy
\begin{document}

\maketitle
\lhead{Preprint}

\begin{abstract}
Latent diffusion models enable high-quality video generation, but video VAE encoding and decoding remain computationally expensive. Existing methods use stage-specific lightweight operators, yet either constrain spatial and temporal computation jointly or remove temporal computation altogether, without distinguishing their compression sensitivities in pretrained Conv3D. To address these limitations, we present LightVAE, a compression framework that exploits redundancy in temporal modeling and network depth. LightVAE exactly reparameterizes eligible causal Conv3D layers into an aggregate spatial response and a difference-driven temporal correction. Based on their distinct compression sensitivities, temporal low-rank compression (TLR) keeps the aggregate spatial kernel dense and applies layer-specific low-rank approximation only to the temporal correction. An algebraically equivalent folded implementation converts the arithmetic savings into practical acceleration. LightVAE further uses width-preserving block pruning to reduce depth and probe-guided feature alignment, guided by feedback from a frozen teacher suffix, to recover quality after compression. The framework also extends to video encoders. Across three video VAE backbones and two GPU platforms, LightVAE consistently accelerates decoding. For the Wan VAE family, it achieves up to $2.24\times$ and $1.11\times$ the decoding throughput of state-of-the-art efficient VAE decoders on NVIDIA H100 and RTX~5090, respectively, with higher DAVIS PSNR and lower LPIPS. On MiniMax-H3, it achieves up to $2.97\times$ the original decoding throughput while retaining 99.51\% of the original model's overall VBench-2.0 score. Our code is available at \url{https://github.com/ModelTC/LightVAE}.
\end{abstract}

\section{Introduction}
\label{sec:introduction}

Recent video generation models have advanced in visual fidelity, motion coherence, and prompt alignment~\citep{yang2024cogvideox,hacohen2024ltxvideo,wan2025wan}. Many operate in a latent space~\citep{rombach2022ldm}, where a video variational autoencoder (VAE) maps between pixels and compact spatiotemporal representations, while a diffusion Transformer (DiT)~\citep{peebles2023dit} generates video latents through iterative denoising. Although this design reduces generation cost, both denoising and VAE inference remain expensive for high-resolution or long videos. Most acceleration methods target denoising. Diffusion distillation reduces the number of sampling steps~\citep{salimans2022progressive,yin2024dmd,yin2024dmd2,yin2025causvid,fan2026phaseddmd}; caching reuses computation across steps~\citep{liu2024teacache}; and efficient attention, quantization, and optimized kernels reduce per-step cost~\citep{zhang2025turbodiffusion,lightx2v}. As denoising becomes faster, VAE inference can account for a larger fraction of end-to-end latency~\citep{zou2025turbovaed,zhu2026flashvaed}. In convolutional video VAEs, causal Conv3D becomes particularly expensive in high-resolution decoder stages, where dense spatiotemporal kernels operate on large feature maps. We therefore focus primarily on decoder compression.

A practical approach is to compress a pretrained decoder while preserving its latent interface and recovering quality through teacher supervision. Turbo-VAED~\citep{zou2025turbovaed} replaces dense Conv3D with depthwise-separable Conv3D in low-resolution stages while retaining dense Conv3D at higher resolutions. Flash-VAED~\citep{zhu2026flashvaed} further combines depthwise-separable Conv3D with channel-pruned Conv2D in high-resolution stages. Flash-VAED's stage-wise replacement experiments show that substituting Conv3D with Conv2D in later, higher-resolution stages yields greater latency reductions with smaller quality losses, providing an empirical basis for reducing temporal computation in these stages. However, the depthwise-separable Conv3D used by both methods constrains joint spatial filtering and channel mixing, limiting its ability to reproduce pretrained dense-convolution responses. Flash-VAED's pruned Conv2D removes local temporal interactions while also narrowing spatial representations, restricting the spatial mappings inherited from the teacher. Conv2D replacement fixes the affected layers to spatial-only processing rather than providing adjustable temporal capacity. These observations motivate examining whether the spatial and temporal contributions of pretrained Conv3D have different compression sensitivities.

Motivated by this question, we introduce \emph{LightVAE}, a compression framework that combines two complementary forms of structural compression: temporal low-rank compression (TLR) within eligible convolutions and width-preserving pruning across network blocks (Figure~\ref{fig:lightvae-overview}). TLR reduces temporal computation within retained layers, whereas block pruning reduces network depth without narrowing the retained feature representations. LightVAE further uses an algebraically equivalent folded implementation for efficient TLR execution and probe-guided feature alignment for reconstruction recovery. 

TLR begins with an exact reparameterization of pretrained causal Conv3D that expresses its output as an aggregate spatial response plus a temporal correction driven by historical--current feature differences. Under the same cumulative singular-value retention criterion, our branch-specific ablations show that temporal-only factorization produces smaller decoder-output errors than spatial-only or joint factorization, while joint-factorization errors remain close to those of spatial-only factorization. These results lead to TLRConv2D, a stateful operator that keeps the aggregate spatial kernel dense and applies truncated SVD only to the temporal correction. Layer-specific ranks control temporal capacity and computational cost; Figure~\ref{fig:temporal-fidelity} illustrates the fidelity benefit of retaining the correction. To translate the reduced arithmetic into wall-clock gains, we derive an algebraically equivalent folded implementation that absorbs the current-frame correction into the spatial kernel and caches low-dimensional history projections. This implementation avoids explicit online feature differences, reduces convolution calls and intermediate tensors without additional rank truncation, and improves measured throughput on both evaluated GPUs.

At the depth level, high shortcut--output cosine similarity suggests potential block redundancy (Figure~\ref{fig:pruning-analysis}). We therefore complement TLR with width-preserving block pruning, adapting TinyFusion's recoverability-aware search~\citep{fang2024tinyfusion} to convolutional and Transformer blocks while retaining required shortcuts, projections, and resampling operations. This depth reduction is complementary to TLR because it acts across blocks rather than within retained temporal convolutions. For reconstruction recovery, uniform feature MSE weights coordinates equally without accounting for their downstream effects. We instead introduce probe-guided feature alignment, which combines gradients from a frozen teacher suffix with signed student--teacher feature discrepancies to construct detached weights. These weights suppress matching directions that locally increase the probe loss in feature space, while the probe loss itself is excluded from the recovery objective.

LightVAE-Pro applies TLR at the original decoder depth, whereas LightVAE-Lite additionally applies width-preserving block pruning using a recoverability-aware search adapted from TinyFusion~\citep{fang2024tinyfusion}. For Transformer-based autoencoders, the framework uses block pruning without TLR. We summarize our contributions as follows:

\begingroup
\setlength{\parskip}{2pt}

\noindent\hangindent=1.4em\hangafter=1
$\blacktriangleright$\hspace{0.3em}
We introduce \emph{TLR}, exploiting the different compression sensitivities of spatial and temporal computation in pretrained causal Conv3D. TLR preserves dense spatial modeling while compressing only the temporal component; an equivalent folded implementation reduces convolution calls and intermediate tensors for practical acceleration.
\par

\noindent\hangindent=1.4em\hangafter=1
$\blacktriangleright$\hspace{0.3em}
We introduce a general structural compression-and-recovery framework that works for both convolutional and Transformer autoencoders. Its probe-guided recovery uses a frozen teacher suffix to weight feature alignment according to how feature discrepancies affect the final reconstruction.
\par

\noindent\hangindent=1.4em\hangafter=1
$\blacktriangleright$\hspace{0.3em}
Across three backbones and two GPUs, LightVAE consistently accelerates decoding. On H100, LightVAE-Lite reaches $2.24\times$ the throughput of Turbo-VAED~\citep{zou2025turbovaed} on Wan2.2; on RTX~5090, it reaches $1.11\times$ that of Flash-VAED~\citep{zhu2026flashvaed} on Wan2.1. In both comparisons, it achieves higher DAVIS PSNR and lower LPIPS. On MiniMax-H3~\citep{minimax2026h3}, it achieves up to $2.97\times$ the original decoding throughput while retaining 99.51\% of the VBench-2.0 overall score.
\par

\endgroup

\section{Related Work}
\label{sec:related-work}

\paragraph{Efficient video autoencoders and operator design.}
Video autoencoder acceleration explores lightweight architectures~\citep{cheng2025leanvae} and compression of pretrained decoders~\citep{zou2025turbovaed,zhu2026flashvaed}, the latter preserving the latent interface of an existing diffusion backbone. Turbo-VAED~\citep{zou2025turbovaed} uses depthwise-separable Conv3D in low-resolution stages and dense Conv3D at higher resolutions. Flash-VAED~\citep{zhu2026flashvaed} combines depthwise-separable Conv3D with channel-pruned Conv2D, demonstrating that later stages tolerate reduced temporal computation. Depthwise-separable Conv3D constrains joint spatial filtering and channel mixing, while channel-pruned Conv2D removes local temporal interactions and narrows spatial representations. Prior work also factorizes spatiotemporal convolutions~\citep{qiu2017p3d,tran2018r2plus1d} or applies low-rank kernel approximation~\citep{lebedev2015speeding}. TLR instead exactly separates an aggregate spatial response from a difference-driven temporal correction, preserving the dense spatial kernel at conversion while controlling temporal capacity through layer-specific ranks.

\paragraph{Structured pruning and knowledge distillation.}
Channel pruning reduces network width~\citep{he2017channelpruning}, whereas block pruning reduces depth while preserving the feature widths of retained blocks. TinyFusion~\citep{fang2024tinyfusion} jointly learns structural selection and parameter adaptation to identify diffusion Transformers with strong post-pruning recoverability. LightVAE adapts this search principle to convolutional and Transformer blocks in video autoencoders. Reconstruction recovery commonly uses output or intermediate-feature distillation~\citep{hinton2015distilling,romero2015fitnets}. CrossKD~\citep{wang2023crosskd} passes student features through a teacher detection head and directly distills the resulting cross-head predictions. Gradient-Guided Knowledge Distillation~\citep{lan2024gkd} uses task gradients to weight task-relevant features during distillation. LightVAE instead uses gradients from a frozen teacher-suffix probe, together with signed student--teacher feature discrepancies, to construct detached feature-alignment weights. These weights suppress matching directions that locally increase the probe loss in feature space. Unlike direct cross-head prediction distillation, the probe loss itself is excluded from the recovery objective.

\section{Preliminaries}
\label{sec:preliminaries}
A video variational autoencoder (VAE) maps a video $\tV$ to a latent representation $\tZ$ through an encoder $\mathcal E$ and reconstructs it as $\tY$ through a decoder $\mathcal D$. Latent video diffusion models generate in this representation space. We focus on decoder compression, but the framework also extends to encoders that use causal Conv3D. Experiments on reference-conditioned encoding are reported in Appendix~\ref{app:h3-ref2v}. Scalars are italic, vectors bold lowercase, matrices bold uppercase, and tensors bold sans-serif uppercase. Subscripts $\mathrm T$, $\mathrm S$, and $\mathrm P$ denote teacher, student, and probe quantities. We use $\operatorname{sg}(\cdot)$ for stop-gradient and $\operatorname{vec}(\cdot)$ for vectorization, omitting batch dimensions throughout. Consider a dense causal Conv3D with $K$ temporal taps and unit temporal stride and dilation:
\begin{equation}
  \tY_t=\sum_{j=0}^{K-1}\tW_j*\tX_{t-j}+\vb,
  \label{eq:causal-conv-tensor}
\end{equation}
where $*$ denotes spatial convolution, $\tX_t\in\mathbb R^{C_{\mathrm{in}}\times H\times W}$, $\tW_j\in\mathbb R^{C_{\mathrm{out}}\times C_{\mathrm{in}}\times k_h\times k_w}$, and $\vb\in\mathbb R^{C_{\mathrm{out}}}$ is broadcast over spatial positions. The output has shape $C_{\mathrm{out}}\times H_{\mathrm{out}}\times W_{\mathrm{out}}$. The index $j=0$ denotes the current frame, independently of kernel storage order, and preceding inputs follow the source layer's temporal padding rule. For the matrix form, let $d=C_{\mathrm{in}}k_hk_w$ and $P=H_{\mathrm{out}}W_{\mathrm{out}}$. Reshape each kernel into $\mW_j\in\mathbb R^{C_{\mathrm{out}}\times d}$ and arrange spatial input patches as columns of $\mX_t\in\mathbb R^{d\times P}$, using the source convolution's spatial stride, dilation, and padding. Then
\begin{equation}
  \mY_t=\sum_{j=0}^{K-1}\mW_j\mX_{t-j}
  +\vb\vone_P^\top,
  \label{eq:causal-conv-matrix}
\end{equation}
where $\mY_t\in\mathbb R^{C_{\mathrm{out}}\times P}$ contains the output features and $\vone_P\in\mathbb R^P$ is the all-ones vector. Patch matrices are used only for derivation and are not materialized during inference.
\section{LightVAE}
\label{sec:lightvae}

\begin{figure}[t]
  \centering
  \includegraphics[width=0.98\linewidth]{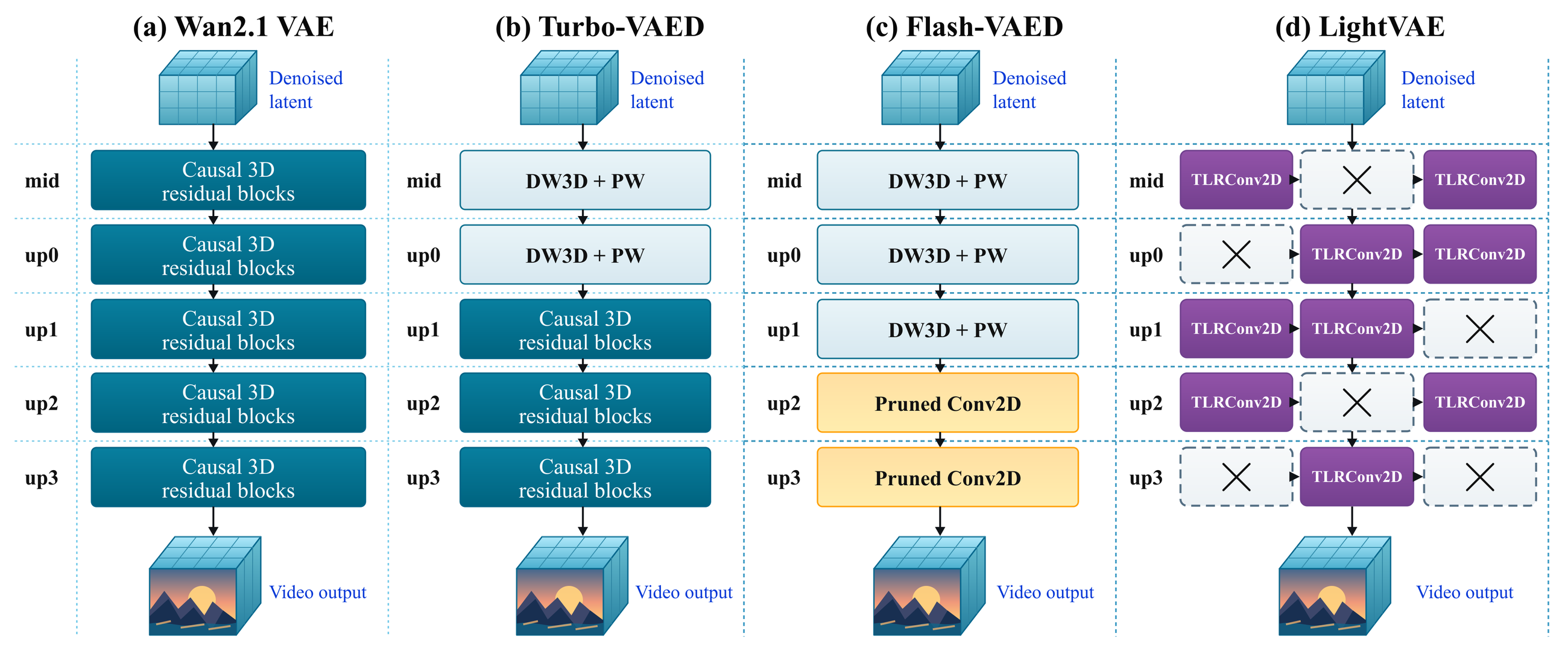}
  \caption{\textbf{Architectural comparison.}
  LightVAE replaces eligible causal Conv3D layers with TLRConv2D
  and optionally removes network blocks.
  TLRConv2D retains the dense aggregate spatial response while applying
  a layer-specific low-rank approximation only to the temporal correction.
    Width-preserving block pruning independently reduces network depth,
    complementing TLR's reduction of per-layer temporal computation.}
  \label{fig:lightvae-overview}
\end{figure}

LightVAE combines temporal low-rank compression with width-preserving block pruning (Figure~\ref{fig:lightvae-overview}). For pruned models, block selection and LoRA merging precede TLR conversion. Probe-guided feature alignment then supports reconstruction recovery, followed by weight folding and projected-history caching for inference.

\subsection{Temporal Low-Rank Compression}
\label{sec:lightvae-tlr}

Temporal low-rank compression (TLR) approximates a pretrained causal convolution while preserving its aggregate spatial response. Below, we derive the resulting TLRConv2D operator. Consider the three-tap convolution in Equation~(\ref{eq:causal-conv-matrix}), with unit temporal stride and dilation.

\paragraph{Spatial response and temporal correction.}
Define the aggregate spatial kernel
$\mS=\mW_0+\mW_1+\mW_2$,
concatenate the historical kernels as
$\mM_{\mathrm{temp}}=[\mW_1,\mW_2]
\in\mathbb R^{C_{\mathrm{out}}\times 2d}$,
and stack the temporal patch differences:
\begin{equation}
  \Delta\mX_t=
  \begin{bmatrix}
    \mX_{t-1}-\mX_t\\
    \mX_{t-2}-\mX_t
  \end{bmatrix}
  \in\mathbb R^{2d\times P}.
  \label{eq:temporal-patch-differences}
\end{equation}
Adding and subtracting the historical kernels applied to the current input yields the exact decomposition
\begin{equation}
  \mY_t=
  \underbrace{\mS\mX_t}_{\text{spatial response}}
  +\underbrace{
    \mM_{\mathrm{temp}}\Delta\mX_t
  }_{\text{temporal correction}}
  +\vb\vone_P^\top.
  \label{eq:spatiotemporal-split}
\end{equation}
The first term applies the sum of all temporal kernel slices to the current feature. The second term corrects this response using differences between the historical and current features.

\paragraph{Approximation allocation.}
TLR retains the aggregate spatial kernel and approximates only the temporal correction. A spatial-kernel error acts on $\mX_t$, whereas a temporal-correction error acts on $\Delta\mX_t$. Preserving $\mS$ therefore removes the spatial approximation term. Temporal continuity motivates this allocation, and the stage-wise comparison in Figure~\ref{fig:wan21-stage-fidelity} evaluates it under the selected truncation rule. The branch-error decomposition and bound are given in Appendix~\ref{app:tlr-analysis}.

\begin{figure*}[t]
  \centering
  \includegraphics[width=0.98\textwidth]{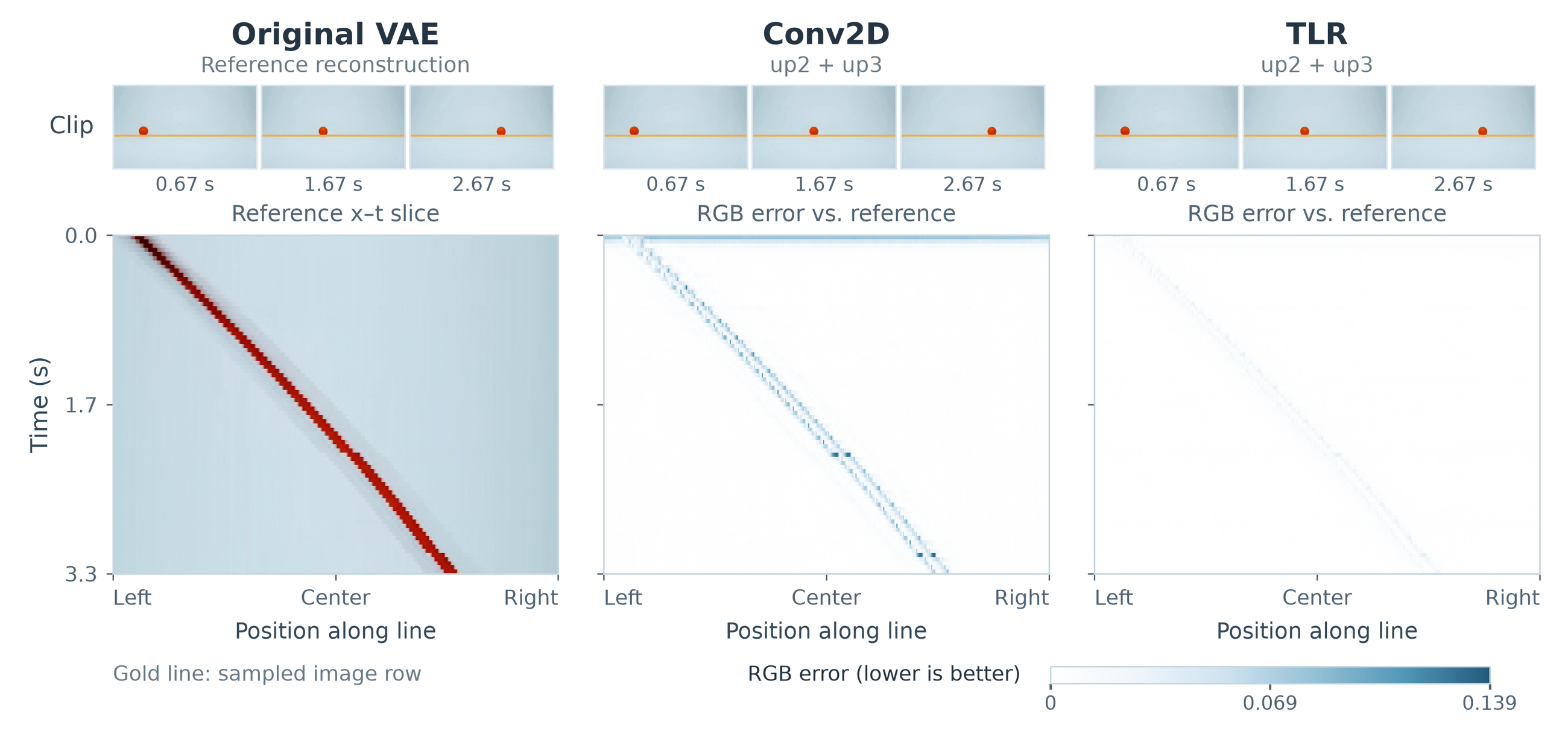}
    \caption{\textbf{Effect of retaining the temporal correction.}
    Only residual convolutions in Wan2.1's \texttt{up2}/\texttt{up3} stages are replaced by Conv2D or TLRConv2D ($\eta=0.6$), without recovery.
    \textbf{Top:} Reconstructed frames with the sampled row marked in gold.
    \textbf{Bottom:} The original VAE's x--t slice and RGB error maps for Conv2D and TLR on a shared scale.}
  \label{fig:temporal-fidelity}
\end{figure*}

\paragraph{Temporal-correction factorization.}
For a prescribed rank $r$, the best rank-$r$ approximation of $\mM_{\mathrm{temp}}$ in the Frobenius norm is given by its truncated singular value decomposition (SVD):
\begin{equation}
\begin{aligned}
  \mM_{\mathrm{temp},r}
  &=\mU_r\mSigma_r\mV_r^\top
   =\mA[\mB_1,\mB_2],\\
  \mA&=\mU_r\mSigma_r,
  \qquad
  [\mB_1,\mB_2]=\mV_r^\top.
\end{aligned}
\label{eq:temporal-svd}
\end{equation}
Here, each $\mB_j\in\mathbb R^{r\times d}$ is implemented as a spatial Conv2D with $r$ output channels, and $\mA\in\mathbb R^{C_{\mathrm{out}}\times r}$ is implemented as a pointwise Conv2D. The first and last $d$ columns of $\mV_r^\top$ form $\mB_1$ and $\mB_2$, respectively. This partition introduces no approximation beyond the SVD and allows the two historical kernels to share the output projection $\mA$. The resulting operator is
\begin{equation}
\begin{aligned}
  \widehat{\mY}_t
  ={}&\mS\mX_t+\vb\vone_P^\top
  +\mA[\mB_1,\mB_2]\Delta\mX_t.
\end{aligned}
\label{eq:tlr}
\end{equation}
The resulting stateful operator, TLRConv2D, implements TLR by imposing a low-rank constraint on $\mM_{\mathrm{temp}}$. At conversion, $\mS$ is preserved exactly, and the temporal factors are initialized from pretrained weights by SVD. The shared threshold $\eta$ specifies the minimum fraction of each layer's temporal singular-value sum to retain, yielding layer-specific ranks according to their singular-value spectra. Layer ranks remain fixed during recovery and inference. For identical layer inputs, SVD initialization gives $\|\widehat{\mY}_t-\mY_t\|_{\mathrm F}\leq\sigma_{r+1}\|\Delta\mX_t\|_{\mathrm F}$. For a fixed truncation, smaller historical--current feature differences yield a smaller bound on the layer-output error. Appendix~\ref{app:tlr-analysis} provides the local error analysis and rank-selection details.
Figure~\ref{fig:temporal-fidelity} compares Conv2D and TLRConv2D replacements applied only to the residual convolutions in the \texttt{up2}/\texttt{up3} stages of the Wan2.1 VAE decoder~\citep{wan2025wan}, without recovery. Relative to the original VAE reconstruction, TLR produces smaller x--t slice errors in this example, illustrating the benefit of retaining the temporal correction.

\paragraph{Computational cost.}
The convolutional multiply--accumulate (MAC) counts per output feature frame are
\begin{equation}
\begin{aligned}
  c_{\mathrm{3D}} &= 3PC_{\mathrm{out}}d,\\
  c_{\mathrm{TLR}} &= P(C_{\mathrm{out}}d+2rd+C_{\mathrm{out}}r).
\end{aligned}
\label{eq:tlr-main-cost}
\end{equation}
TLR reduces the MAC count when $r<2C_{\mathrm{out}}d/(2d+C_{\mathrm{out}})$. Appendix~\ref{app:tlr-cost} provides the derivation and counting conventions.

\subsection{Weight-Folded Execution with Projected-History Caching}
\label{sec:lightvae-efficient}

\begin{figure*}[t]
  \centering
  \includegraphics[width=0.98\textwidth]{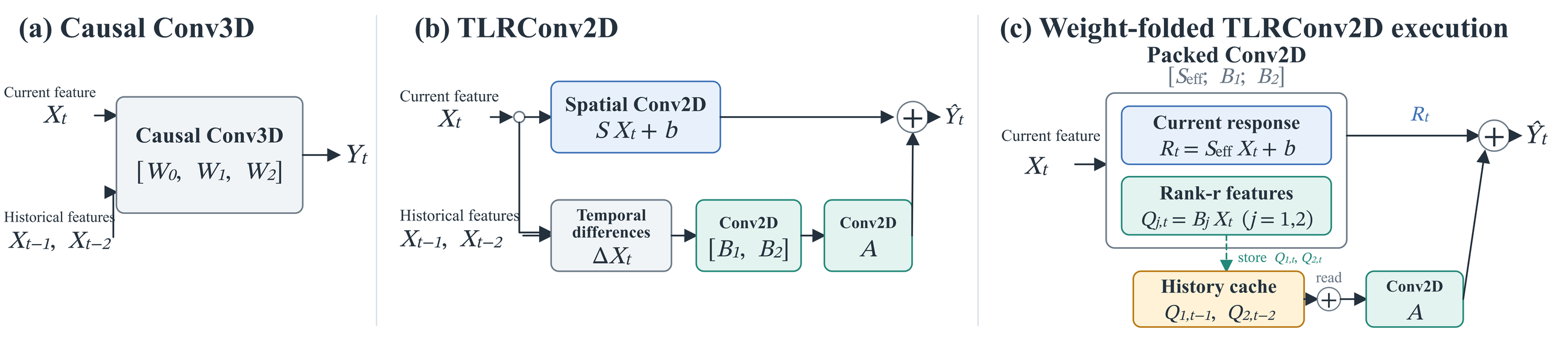}
  \caption{Causal Conv3D, TLRConv2D, and its weight-folded execution.}
  \label{fig:tlr-execution}
\end{figure*}

The difference form in Equation~(\ref{eq:tlr}) requires three spatial-convolution calls and explicitly constructs two temporal differences. As illustrated in Figure~\ref{fig:tlr-execution}, we reduce operator-call and intermediate-tensor overhead by folding the current-frame correction into the spatial kernel and computing the current response and low-rank features from the same current-frame input.

\paragraph{Weight folding.}
After recovery, the fixed parameters allow the current-frame terms to be collected as
\begin{equation}
  \mS_{\mathrm{eff}}=\mS-\mA(\mB_1+\mB_2).
  \label{eq:tlr-folded-kernel}
\end{equation}
The product contracts the intermediate channel dimension and is reshaped to the same spatial-kernel dimensions as $\mS$. For each incoming frame, define
\begin{equation}
  \mathbf Q_{j,t}=\mB_j\mX_t
  \in\mathbb R^{r\times P},
  \qquad j\in\{1,2\}.
  \label{eq:tlr-history-projection}
\end{equation}
Equation~(\ref{eq:tlr}) then becomes
\begin{equation}
  \widehat{\mY}_t=
  \underbrace{
    \mS_{\mathrm{eff}}\mX_t+\vb\vone_P^\top
  }_{\mathbf R_t}
  +\mA(\mathbf Q_{1,t-1}+\mathbf Q_{2,t-2}).
  \label{eq:tlr-folded-forward}
\end{equation}
The current-frame correction is contained in $\mathbf R_t$, leaving only the lagged projections in the history term.

\paragraph{Joint convolution and projected history.}
Because $\mS_{\mathrm{eff}}$, $\mB_1$, and $\mB_2$ share the same input, their output channels can be concatenated into one Conv2D:
\begin{equation}
  \begin{bmatrix}
    \mathbf R_t\\
    \mathbf Q_{1,t}\\
    \mathbf Q_{2,t}
  \end{bmatrix}
  =
  \begin{bmatrix}
    \mS_{\mathrm{eff}}\\
    \mB_1\\
    \mB_2
  \end{bmatrix}\mX_t
  +
  \begin{bmatrix}
    \vb\\
    \mathbf 0_r\\
    \mathbf 0_r
  \end{bmatrix}\vone_P^\top.
  \label{eq:tlr-joint-convolution}
\end{equation}
This joint convolution has $C_{\mathrm{out}}+2r$ output channels and retains the source spatial stride, dilation, and padding. The cached projections $\mathbf Q_{1,t-1}$ and $\mathbf Q_{2,t-2}$ are summed, projected by $\mA$, and added to $\mathbf R_t$. The current projections are cached for outputs at $t+1$ and $t+2$, respectively, with the required history read before any cache entry is overwritten. Cache initialization follows the source temporal-padding rule; the state persists across chunks of the same video and is reset between videos. Folding preserves the MAC count in Equation~\ref{eq:tlr-main-cost} while reducing convolution calls and intermediate tensors. 

\subsection{Block Selection and Recovery}
\label{sec:lightvae-search}
\label{sec:lightvae-recovery}

\paragraph{Block selection.}
We adapt TinyFusion's recoverability-aware search~\citep{fang2024tinyfusion} to video VAE blocks under fixed group-wise retention quotas. Required shortcuts, projections, and resampling operations are retained. Selection logits and shared LoRA parameters~\citep{hu2021lora} are trained with frozen base weights using an $\mathcal L_1$ reconstruction term and LPIPS~\citep{zhang2018unreasonable}. After search, LoRA updates are merged into the retained weights before TLRConv2D initialization. The search applies to convolutional and compatible Transformer blocks. Appendix~\ref{app:search} summarizes the grouped search, pruning units, and export procedure.

\paragraph{Probe-guided recovery.}
At a compatible feature boundary, a frozen teacher suffix maps the student's feature to a reconstruction, which is compared with the detached teacher output. Let $\mathbf g$ denote the probe loss's input gradient and $\boldsymbol\delta$ the signed student--teacher feature discrepancy. Ordinary feature MSE updates coordinate $i$ along $-\delta_i$, which locally increases the probe loss when $\delta_i g_i<0$. We use $\mathbf g$ and $\boldsymbol\delta$ to construct nonnegative alignment weights $w_i$ that suppress these conflicting updates. The weighted feature loss is $\mathcal L_{\mathrm{feature}}=N_f^{-1}\sum_i\operatorname{sg}(w_i)\delta_i^2$, where $N_f$ is the number of feature elements. Appendix~\ref{app:probe} defines the weighting rule and analyzes its local feature-space alignment. The student is trained with
\begin{equation}
\begin{aligned}
  \mathcal L_{\mathrm{recon}}={}&
  \mathcal L_1(\tY_{\mathrm S},\tV)
  +\lambda_{\mathrm{perc}}
    \operatorname{LPIPS}(\tY_{\mathrm S},\tV)
  +\lambda_{\mathrm{adv}}\mathcal L_{\mathrm{adv}},\\
  \mathcal L_{\mathrm{restore}}={}&
  \mathcal L_{\mathrm{recon}}
  +\lambda_{\mathrm F}\mathcal L_{\mathrm{feature}}.
\end{aligned}
\label{eq:recovery-loss}
\end{equation}
Here, $\mathcal{L}_1$, LPIPS, and $\mathcal{L}_{\mathrm{adv}}$ denote reconstruction, perceptual, and adversarial losses. Loss coefficients and training settings are given in Appendix~\ref{app:training-settings}. We cycle through compatible group boundaries, evaluating one probe per iteration. Its gradients determine detached feature weights; its loss is excluded from $\mathcal{L}_{\mathrm{restore}}$. Teacher and probe computations are used only during recovery.
\section{Experiments}
\label{sec:experiments}

\subsection{Experimental Setup}
\label{sec:experimental-setup}

For Wan2.1/2.2~\citep{wan2025wan,wan2025wan22}, Pro retains all 14 residual blocks, while Lite keeps one residual main branch per stage ($14\!\rightarrow\!5$). MiniMax-H3~\citep{minimax2026h3} uses block pruning only, reducing decoder Transformer blocks ($36\!\rightarrow\!12$) and encoder residual main branches ($12\!\rightarrow\!6$). Throughput is measured on NVIDIA H100 and GeForce RTX~5090 using 81-frame $480\times832$/$704\times1280$ clips for Wan2.1/Wan2.2 and 124-frame $768\times1344$ clips for MiniMax-H3. Reconstruction uses 33 consecutive native-resolution frames on UCF-101~\citep{soomro2012ucf101} and all frames at $720\times1280$ on the DAVIS 2019 Challenge test set~\citep{caelles2019davis}. Where required, last-frame repetition yields $4n+1$ inputs; repeated frames are excluded from scoring. Ablations use 33-frame $480\times832$ clips unless specified otherwise. Video-level RGB PSNR and frame-averaged SSIM~\citep{wang2004ssim} and LPIPS-VGG~\citep{zhang2018unreasonable,simonyan2015very} are averaged equally across videos. For VBench-2.0~\citep{zheng2025vbench2}, LightX2V~\citep{lightx2v} uses the throughput-test resolutions with 81/121/124 frames for Wan2.1/Wan2.2/MiniMax-H3, respectively. Within each backbone, only the decoder varies; prompts, seeds, and sampling settings remain fixed. Appendix~\ref{app:training-settings} provides compression hyperparameters, training settings, loss coefficients, and software versions.

\subsection{Main Results}
\label{sec:exp-wan}
\label{sec:exp-generation}

\paragraph{Reconstruction and generation quality.}
Tables~\ref{tab:main-results} and~\ref{tab:vbench2-results} summarize LightVAE's speed--quality trade-offs. On Wan2.1, Lite achieves $1.12\times$/$1.11\times$ the throughput of Flash-VAED on H100/RTX~5090, with higher DAVIS PSNR and lower LPIPS, while Pro closely preserves the original reconstruction quality. On Wan2.2, Pro surpasses Turbo-VAED on all reconstruction metrics across both datasets while achieving higher H100 throughput (38.84 vs.\  29.24 FPS). Across the Wan VAEs, Lite reaches up to $4.06\times$/$4.20\times$ the original throughput on H100/RTX~5090. On RTX~5090, Wan2.2 Lite reaches 31.87 FPS, whereas Turbo-VAED runs out of memory under our 81-frame protocol, compared with the standard 17-frame evaluation in its original paper~\citep{zou2025turbovaed}. Both LightVAE variants use $r=0$ in the converted \texttt{up2}/\texttt{up3} residual convolutions, consistent with evidence that later stages tolerate reduced temporal computation~\citep{zhu2026flashvaed}. The ablations in Section~\ref{sec:ablations} separately show that TLR gives the lowest matched-MAC error among the tested allocations, block pruning reaches higher recovered PSNR than channel pruning at similar pre-recovery throughput, probe-guided recovery improves over reconstruction-only recovery, and folding increases throughput on both GPUs. On MiniMax-H3, Lite achieves $2.95\times$/$2.97\times$ the original decoding throughput on H100/RTX~5090 while retaining 99.51\% of the original VBench-2.0 overall score. Reference encoding is accelerated by up to $2.68\times$, with jointly compressed encoders and decoders producing visually similar outputs. Appendices~\ref{app:vbench2-breakdown} and~\ref{app:h3-ref2v} provide detailed scores and reference-conditioned comparisons.

\begin{table}[t]
\centering
\caption{VAE reconstruction quality and decoding throughput on NVIDIA H100 and RTX~5090. OOM denotes out of memory. Best compressed results per backbone are \textbf{bold}.}
\label{tab:main-results}
\begingroup
\small
\setlength{\dashlinedash}{2pt}
\setlength{\dashlinegap}{2pt}
\setlength{\tabcolsep}{4pt}
\renewcommand{\arraystretch}{1.1}
\resizebox{\linewidth}{!}{%
\begin{tabular}{@{}lcccccccc@{}}
\toprule
\multirow{2}{*}{Model}
& \multicolumn{2}{c}{FPS$\uparrow$}
& \multicolumn{3}{c}{UCF-101}
& \multicolumn{3}{c}{DAVIS-2019} \\
\cmidrule(lr){2-3}
\cmidrule(lr){4-6}
\cmidrule(l){7-9}
& NVIDIA H100 & RTX~5090
& PSNR$\uparrow$ & SSIM$\uparrow$ & LPIPS$\downarrow$
& PSNR$\uparrow$ & SSIM$\uparrow$ & LPIPS$\downarrow$ \\
\midrule
Wan2.1 VAE
& 52.48 & 30.92
& 33.5021 & 0.94087 & 0.05343
& 34.2021 & 0.92363 & 0.08627 \\
\hdashline

Flash-VAED
& 191.04 & 116.55
& 30.7241 & 0.91399 & 0.09013
& 31.9097 & 0.90491 & 0.12768 \\

LightVAE-Pro
& 104.02 & 61.88
& \textbf{33.4373} & \textbf{0.94339} & \textbf{0.05585}
& \textbf{33.9764} & \textbf{0.92071} & \textbf{0.09478} \\

LightVAE-Lite
& \textbf{213.28} & \textbf{129.94}
& 30.9058 & 0.91453 & 0.09027
& 32.4032 & 0.90302 & 0.11575 \\

\midrule

Wan2.2 VAE
& 23.95 & 10.18
& 33.9193 & 0.94532 & 0.05134
& 34.2463 & 0.92134 & 0.08315 \\
\hdashline

Turbo-VAED
& 29.24 & OOM
& 31.8020 & 0.93116 & 0.06627
& 30.0160 & 0.89384 & 0.13801 \\

LightVAE-Pro
& 38.84 & 18.58
& \textbf{33.3065} & \textbf{0.94110} & \textbf{0.05776}
& \textbf{33.5184} & \textbf{0.91585} & \textbf{0.09774} \\

LightVAE-Lite
& \textbf{65.41} & \textbf{31.87}
& 30.8672 & 0.91330 & 0.08969
& 31.5560 & 0.89150 & 0.13570 \\

\midrule

MiniMax-H3 VAE
& 18.15 & 9.85
& 31.6361 & 0.91876 & 0.09328
& 32.2907 & 0.89980 & 0.14918 \\
\hdashline

LightVAE-Lite
& \textbf{53.52} & \textbf{29.23}
& \textbf{28.8349} & \textbf{0.87828} & \textbf{0.13696}
& \textbf{29.3048} & \textbf{0.85393} & \textbf{0.21310} \\

\bottomrule
\end{tabular}%
}
\endgroup
\end{table}

\begin{table*}[t]
\centering
\caption{VBench-2.0 generation quality (0--100; higher is better).}
\label{tab:vbench2-results}
\begingroup
\small
\setlength{\dashlinedash}{2pt}
\setlength{\dashlinegap}{2pt}
\setlength{\tabcolsep}{4pt}
\renewcommand{\arraystretch}{1.1}
\resizebox{\textwidth}{!}{%
\begin{tabular}{@{}llcccccc@{}}
\toprule
\multirow{2}{*}{Generation backbone}
& \multirow{2}{*}{Decoder}
& \multicolumn{6}{c}{VBench-2.0$\uparrow$} \\
\cmidrule(l){3-8}
& & \shortstack{Human\\Fidelity}
& Controllability
& Creativity
& Physics
& Commonsense
& Overall \\
\midrule

\multirow{4}{*}{Wan2.1-1.3B}
& Original decoder
& 85.3388 & 21.7736 & 52.3006
& 43.8093 & 61.7110 & 52.9867 \\
\cdashline{2-8}
& Flash-VAED
& 83.1481 & 21.4708 & 49.6575
& \textbf{46.7925} & 61.4237 & 52.4985 \\
& LightVAE-Pro
& \textbf{84.8408} & \textbf{21.8701} & \textbf{51.9213}
& 45.8570 & 61.1363 & \textbf{53.1251} \\
& LightVAE-Lite
& 84.1943 & 21.8338 & 48.4021
& 44.8614 & \textbf{61.7110} & 52.2005 \\
\midrule

\multirow{4}{*}{Wan2.2-5B}
& Original decoder
& 86.3657 & 22.6547 & 56.6134
& 53.5895 & 65.1593 & 56.8765 \\
\cdashline{2-8}
& Turbo-VAED
& 86.3387 & \textbf{23.2704} & \textbf{58.1631}
& 54.8722 & \textbf{66.0214} & \textbf{57.7332} \\
& LightVAE-Pro
& \textbf{86.4226} & 23.0824 & 57.6016
& \textbf{55.3117} & 65.1593 & 57.5155 \\
& LightVAE-Lite
& 85.9445 & 23.1687 & 55.7466
& 54.6638 & 65.1593 & 56.9366 \\
\midrule

\multirow{2}{*}{MiniMax-H3}
& Original decoder
& 83.9687 & 54.4288 & 45.6439
& 78.3034 & 72.0710 & 66.8832 \\
\cdashline{2-8}
& LightVAE-Lite
& \textbf{82.6056} & \textbf{55.2702} & \textbf{45.4439}
& \textbf{75.3712} & \textbf{74.0825} & \textbf{66.5546} \\

\bottomrule
\end{tabular}%
}
\endgroup
\end{table*}

\subsection{Ablation Studies}
\label{sec:ablations}

\begin{figure}[!htbp]
  \centering
  \edef\LightVAEPanelHeight{\the\dimexpr0.33\textwidth\relax}
  \captionsetup{font=small,labelfont=normalfont,textfont=normalfont,
    justification=raggedright,singlelinecheck=false,skip=3pt}
  \begin{minipage}[t]{0.485\textwidth}
    \vspace{0pt}\centering
    \parbox[c][\LightVAEPanelHeight][c]{\linewidth}{%
      \centering
      \includegraphics[width=\linewidth,height=\LightVAEPanelHeight,
        keepaspectratio]{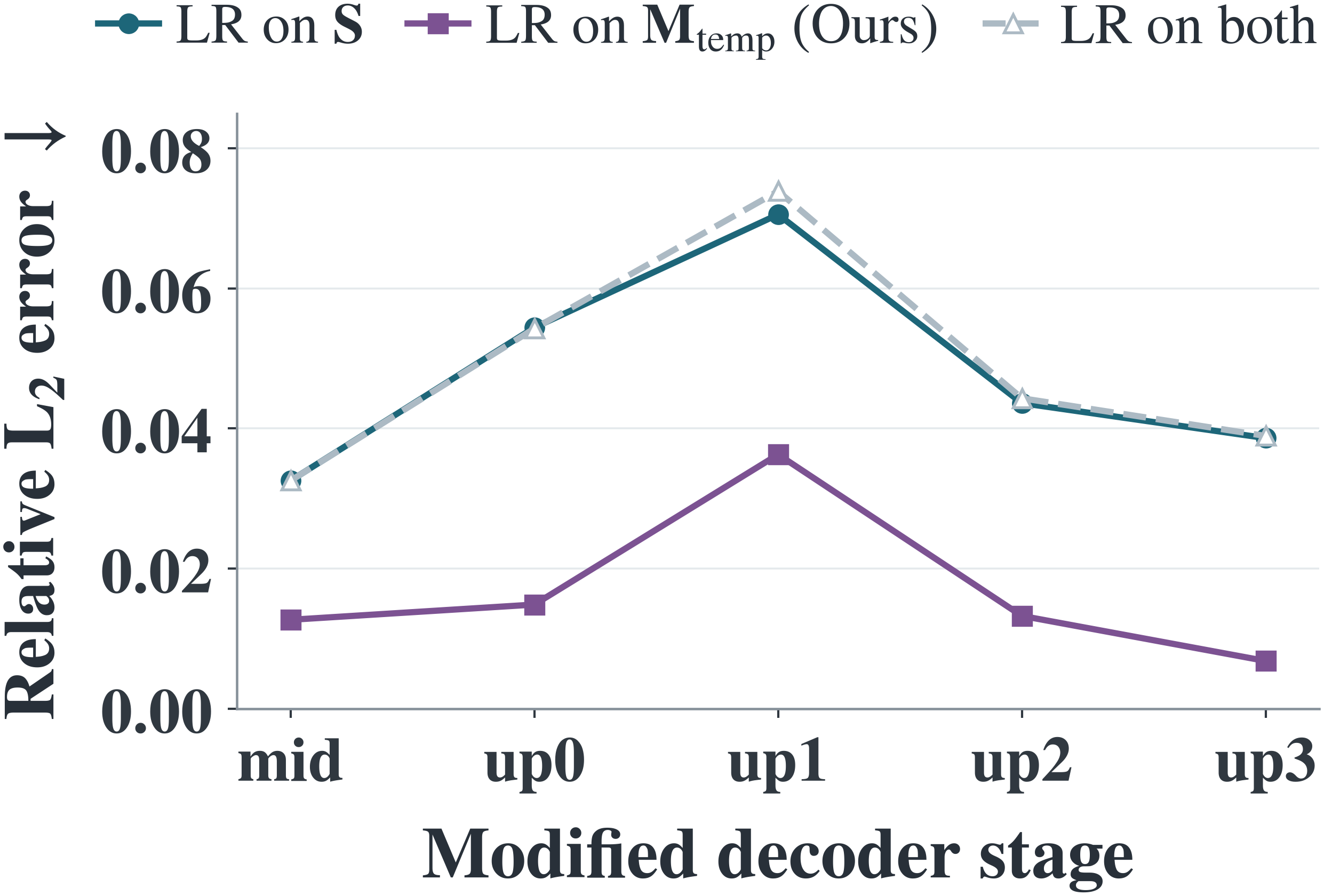}%
    }
    \caption{Mean relative $L_2$ error under stage-wise approximation.}
    \label{fig:wan21-stage-fidelity}
    \label{fig:wan21-branch-fidelity}
  \end{minipage}\hfill
  \begin{minipage}[t]{0.485\textwidth}
    \vspace{0pt}\centering
    \parbox[c][\LightVAEPanelHeight][c]{\linewidth}{%
      \centering
      \includegraphics[width=\linewidth,height=\LightVAEPanelHeight,
        keepaspectratio]{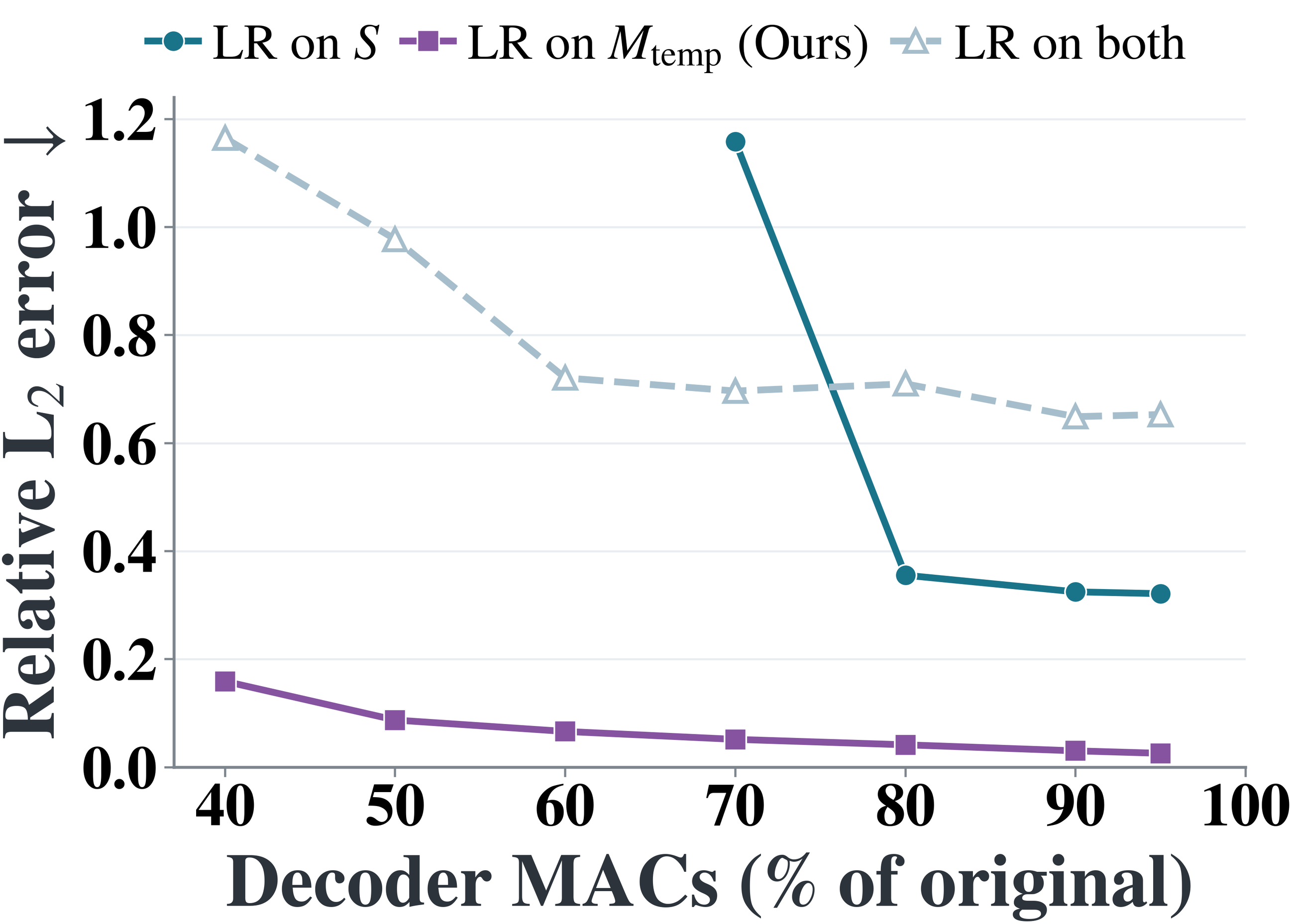}%
    }
    \caption{Mean relative $L_2$ error at matched MAC budgets.}
    \label{fig:matched-mac}
  \end{minipage}
\end{figure}

\begin{figure}[!htbp]
  \centering
  \edef\LightVAEPanelHeight{\the\dimexpr0.33\textwidth\relax}
  \captionsetup{font=small,labelfont=normalfont,textfont=normalfont,
    justification=raggedright,singlelinecheck=false,skip=3pt}
  \begin{minipage}[t]{0.48\textwidth}
    \vspace{0pt}\centering
    \parbox[c][\LightVAEPanelHeight][c]{\linewidth}{%
      \centering
      \includegraphics[width=\linewidth,height=\LightVAEPanelHeight,
        keepaspectratio]{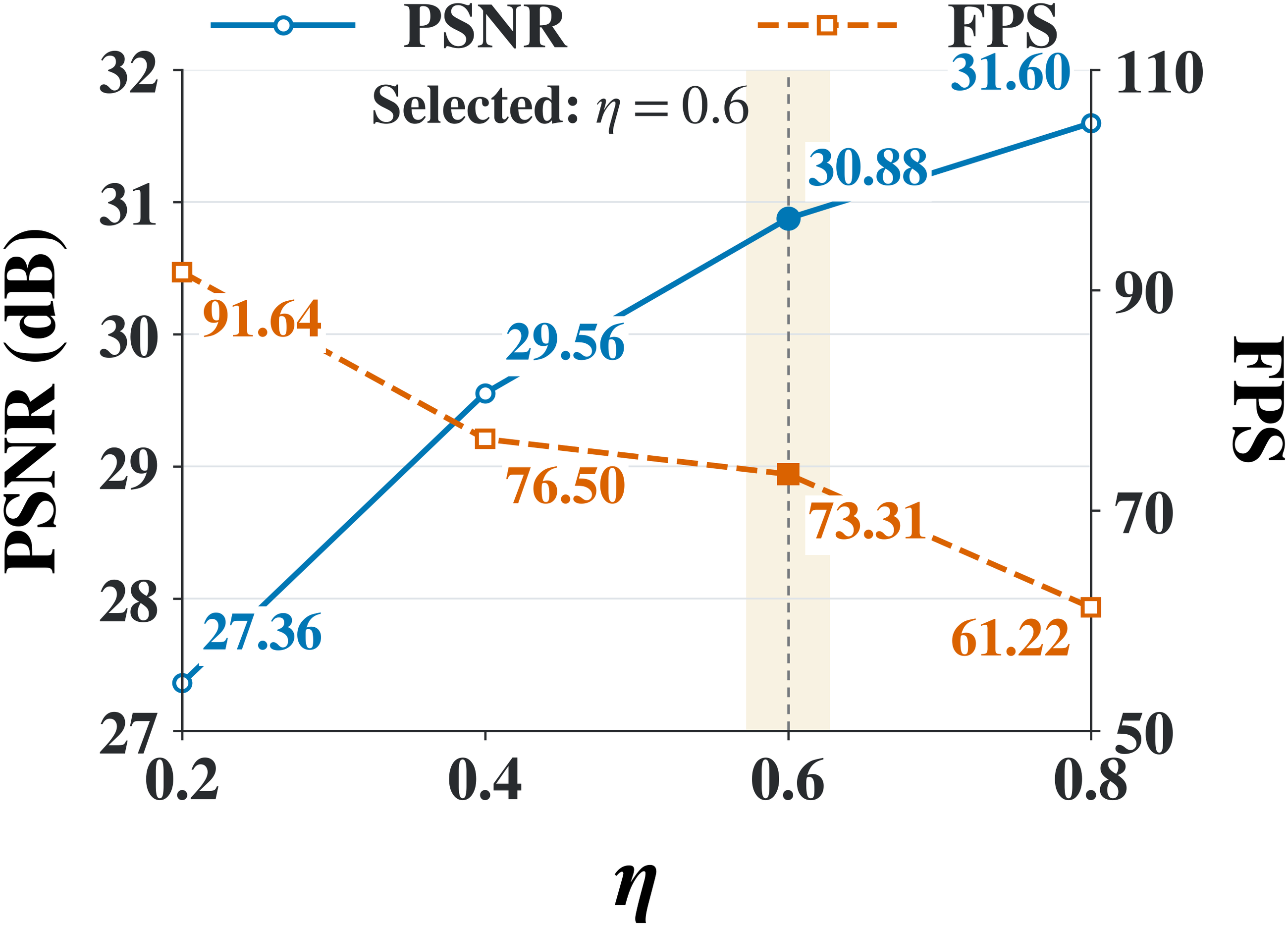}%
    }
    \caption{PSNR and throughput versus retention threshold $\eta$.}
    \label{fig:rank-ablation}
  \end{minipage}\hfill
  \begin{minipage}[t]{0.48\textwidth}
    \vspace{0pt}\centering
    \parbox[c][\LightVAEPanelHeight][c]{\linewidth}{%
      \centering
      \includegraphics[width=\linewidth,height=\LightVAEPanelHeight,
        keepaspectratio]{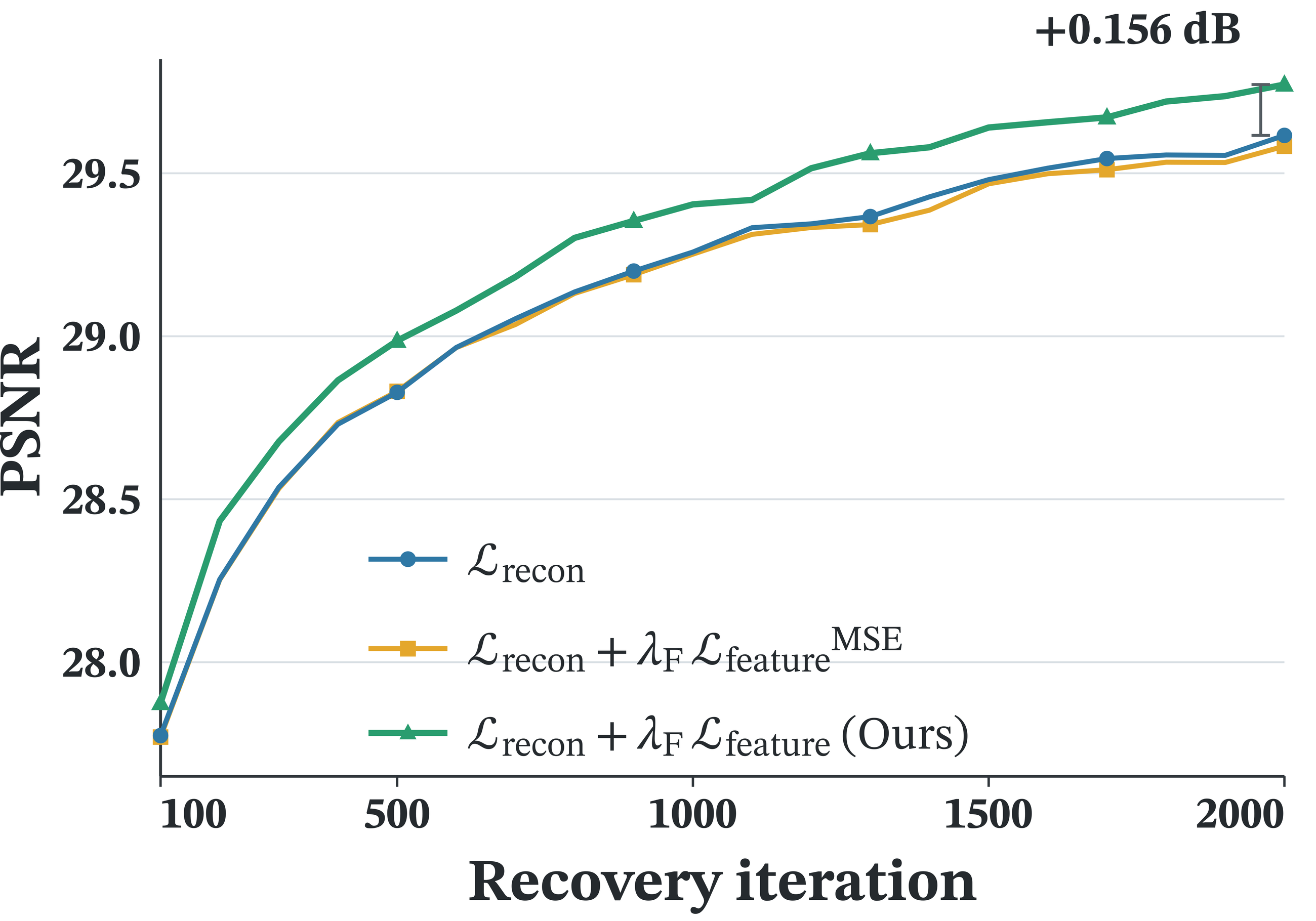}%
    }
    \caption{DAVIS PSNR with different recovery losses.}
    \label{fig:recovery-ablation}
  \end{minipage}
\end{figure}

\paragraph{Spatial--temporal approximation and rank selection.}
At a shared retention threshold, temporal-only approximation yields the smallest output errors, while joint errors closely track spatial-only errors (Figure~\ref{fig:wan21-stage-fidelity}). Temporal errors decrease from \texttt{up1} to \texttt{up3}, consistent with Flash-VAED's stage-wise findings~\citep{zhu2026flashvaed}. TLR also achieves the lowest error among the tested matched-MAC allocations without recovery (Figure~\ref{fig:matched-mac}). Near its 70\% MAC floor, spatial-only compression leaves little spatial capacity; joint compression allocates more budget to $S$ and reduces error. At 80\%--95\%, spatial-only compression regains spatial capacity while retaining dense $M_{\mathrm{temp}}$, outperforming the joint allocation that keeps both branches low-rank. These results support preserving $S$ and prioritizing compression of $M_{\mathrm{temp}}$. Appendix~\ref{app:matched-mac} details the protocols and allocation constraints. Figure~\ref{fig:rank-ablation} shows increasing fidelity with higher $\eta$ at reduced throughput; we use $\eta=0.6$ as the default balance.

\begin{figure*}[t]
  \centering
  \includegraphics[width=\textwidth]{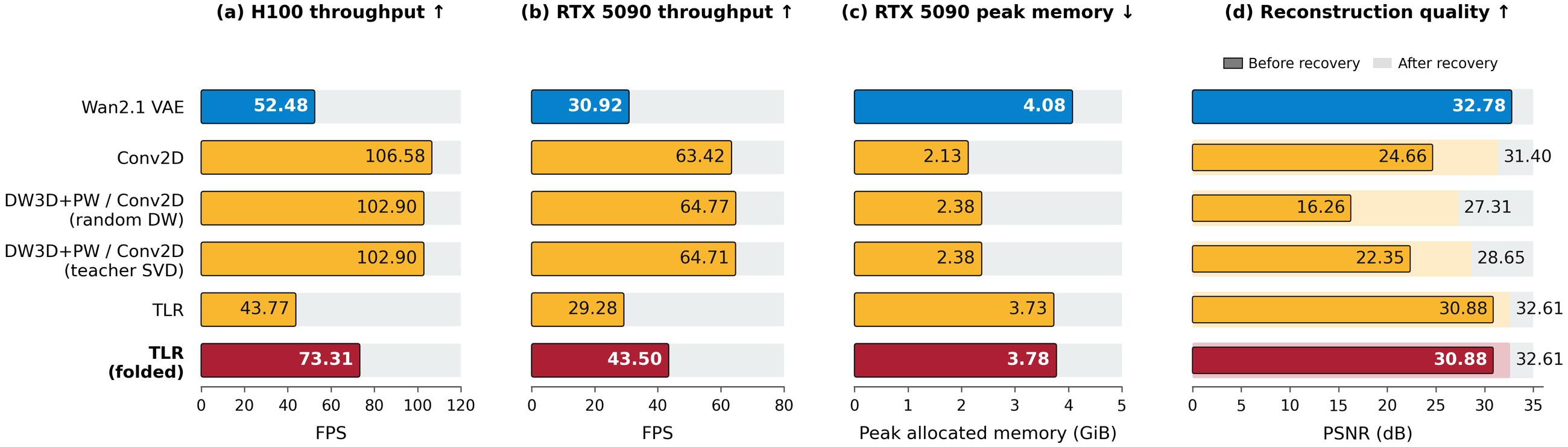}
    \caption{Wan2.1 VAE operator comparison at the original depth and width.
    (a,b) H100/RTX~5090 throughput; (c) RTX~5090 peak allocated memory;
    (d) DAVIS PSNR before (dark) and after (light) matched recovery.}
  \label{fig:operator-ablation}
\end{figure*}

\paragraph{Operator quality and inference efficiency.}
Figure~\ref{fig:operator-ablation} compares TLR with summed-slice Conv2D and DW3D+PW/Conv2D hybrids using Flash-VAED's layout~\citep{zhu2026flashvaed}. Teacher-SVD initialization, adapted from network decoupling~\citep{guo2018nd}, improves the hybrids over random-DW initialization, but TLR retains higher PSNR before and after matched recovery. This supports preserving dense spatial mappings while approximating the temporal correction. Folding improves throughput on both GPUs. Retaining these mappings and the low-rank correction costs more than the lighter operators, while combining TLR with block pruning provides the favorable speed--quality trade-offs in Table~\ref{tab:main-results}.

\begin{table}[t]
\centering
\caption{Decoding throughput (FPS$\uparrow$) under individual and joint
compression. Parentheses show speedup over the original decoder on
the same GPU and backbone.}
\label{tab:compression-complementarity}
\begingroup
\footnotesize
\setlength{\tabcolsep}{4pt}
\renewcommand{\arraystretch}{1.18}
\begin{tabular}{@{}ccc@{\hspace{10pt}}cccc@{}}
\toprule
\multicolumn{3}{c}{Compression}
& \multicolumn{2}{c}{NVIDIA H100}
& \multicolumn{2}{c}{RTX~5090} \\
\cmidrule(lr){1-3}
\cmidrule(lr){4-5}
\cmidrule(l){6-7}
\shortstack{Depth\\pruning} & TLR
& \shortstack{$r=0$\\up2/up3}
& Wan2.1 & Wan2.2 & Wan2.1 & Wan2.2 \\
\midrule
$\times$ & $\times$ & $\times$
& 52.48 ($1.00\times$) & 23.95 ($1.00\times$)
& 30.92 ($1.00\times$) & 10.18 ($1.00\times$) \\
\midrule
$\checkmark$ & $\times$ & $\times$
& 123.20 ($2.35\times$) & 44.19 ($1.85\times$)
& 71.57 ($2.31\times$) & 19.73 ($1.94\times$) \\

$\times$ & $\checkmark$ & $\times$
& 73.31 ($1.40\times$) & 27.24 ($1.14\times$)
& 43.50 ($1.41\times$) & 12.83 ($1.26\times$) \\

$\checkmark$ & $\checkmark$ & $\times$
& 165.54 ($3.15\times$) & 51.65 ($2.16\times$)
& 100.08 ($3.24\times$) & 24.51 ($2.41\times$) \\

$\checkmark$ & $\checkmark$ & $\checkmark$
& \textbf{213.28} ($\mathbf{4.06}\times$)
& \textbf{65.41} ($\mathbf{2.73}\times$)
& \textbf{129.94} ($\mathbf{4.20}\times$)
& \textbf{31.87} ($\mathbf{3.13}\times$) \\
\bottomrule
\end{tabular}
\endgroup
\end{table}

\paragraph{Joint operator and depth compression.}
Table~\ref{tab:compression-complementarity} uses the Wan input settings of Table~\ref{tab:main-results}. Pruning retains 5 of 14 residual main branches, while TLR uses $\eta=0.6$ throughout. Depth pruning alone achieves $1.85\times$--$2.35\times$ the original throughput, and folded TLR alone achieves $1.14\times$--$1.41\times$. Crucially, applying folded TLR after pruning provides a further $1.17\times$--$1.40\times$ throughput increase over pruning alone. This incremental gain demonstrates their complementarity: pruning reduces network depth, whereas TLR reduces the temporal-convolution cost within the retained network. LightVAE-Lite combines depth pruning with TLR, setting $r=0$ in \texttt{up2}/\texttt{up3} and using $\eta=0.6$ in the remaining stages. Compared with applying $\eta=0.6$ in all stages, this configuration improves decoding throughput by $1.27\times$--$1.30\times$, consistent with the LightVAE-Lite results in Table~\ref{tab:main-results}.

\begin{figure}[!htbp]
  \centering
  \captionsetup{font=small,labelfont=normalfont,textfont=normalfont,
    justification=raggedright,singlelinecheck=false,skip=3pt}
  \includegraphics[width=0.96\linewidth]{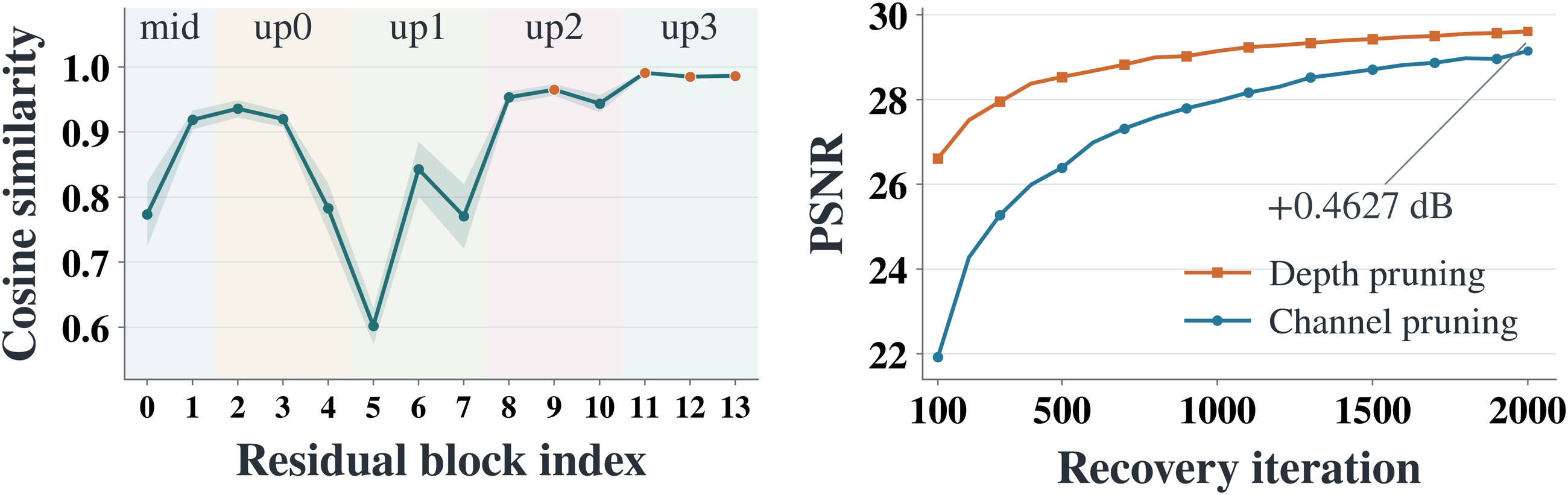}
  \caption{Shortcut--output cosine similarity (left) and DAVIS PSNR
    under depth and channel pruning (right).}
  \label{fig:pruning-analysis}
\end{figure}

\paragraph{Block pruning and recovery.}
High shortcut--output similarity suggests depth redundancy (Figure~\ref{fig:pruning-analysis}, left; Appendix~\ref{app:pruning-feature-visualization}). For comparable inference cost, block pruning retains 6 of 14 residual main branches, while channel pruning removes approximately 41.7\% of channels. Both retain dense Conv3D, reaching 45.58/42.69 FPS on RTX~5090 before recovery. Block pruning converges faster and reaches higher PSNR under matched recovery (right), supporting width-preserving depth reduction. Separately, Figure~\ref{fig:recovery-ablation} fixes the architecture at 5 of 14 branches: uniform feature MSE underperforms reconstruction alone, whereas probe-guided alignment improves recovery. Appendix~\ref{app:probe} analyzes the local alignment of feature updates with the probe objective.

\section{Conclusion}
We presented LightVAE, combining temporal low-rank compression, folded execution, width-preserving block pruning, and probe-guided recovery. It achieves up to $2.24\times$/$1.11\times$ the throughput of evaluated efficient Wan decoders on H100/RTX~5090 with higher DAVIS PSNR and lower LPIPS. On MiniMax-H3, block pruning provides up to $2.97\times$ decoding throughput while retaining 99.51\% of the VBench-2.0 score, and extends to efficient reference encoding.

\subsection*{AI use statement}
In this work, generative AI tools were used solely to improve the language, readability, and presentation of the manuscript. They were not used for any task requiring disclosure under the ICLR 2027 Policy on the Use of AI by Authors. The authors reviewed all AI-assisted revisions to ensure that the original technical meaning was preserved and take full responsibility for the final content of the manuscript, including all AI-assisted language edits.
\FloatBarrier
\bibliography{iclr2027_conference}
\bibliographystyle{iclr2027_conference}
\clearpage
\appendix
\clearpage
\appendix

\addtocontents{toc}{\protect\setcounter{tocdepth}{2}}
\begingroup
    \setcounter{tocdepth}{-10}
    \renewcommand{\contentsname}{Appendix}
    \tableofcontents
\endgroup
\medskip

\numberwithin{equation}{section}
\numberwithin{figure}{section}
\numberwithin{table}{section}

\section{Limitations}
\label{app:limitations}

The convolutional evaluation of TLR currently focuses on the Wan VAE family; the MiniMax-H3 experiments evaluate block pruning rather than TLR. Temporal ranks are selected from weight spectra, block retention follows predefined group-wise quotas, and measured latency depends on hardware and implementation. In addition, the probe weighting provides a local feature-space alignment property rather than a parameter-space descent guarantee. Future work could learn ranks and retention counts jointly from activation and latency statistics and evaluate TLR across additional video-autoencoder families.

\section{Analysis of Temporal Low-Rank Compression}
\label{app:tlr-analysis}

Unless otherwise stated, we analyze a three-tap dense causal convolution with unit temporal stride and unit dilation, identical layer inputs, and identical bias and boundary handling. The SVD error results describe local conversion at initialization, whereas the folding identity in Appendix~\ref{app:tlr-folding} remains exact after recovery. For LightVAE-Lite, the source weights include the LoRA updates merged after block selection.

\subsection{Local Error and Comparison with Conv2D}
\label{app:tlr-error-details}

\paragraph{Exact decomposition and branch errors.}
Substituting $\mS=\mW_0+\mW_1+\mW_2$ into
Equation~(\ref{eq:spatiotemporal-split}) gives
\begin{equation}
\begin{aligned}
  &\mS\mX_t+\mM_{\mathrm{temp}}\Delta\mX_t\\
  &=(\mW_0+\mW_1+\mW_2)\mX_t
    +\mW_1(\mX_{t-1}-\mX_t)
    +\mW_2(\mX_{t-2}-\mX_t)\\
  &=\mW_0\mX_t+\mW_1\mX_{t-1}+\mW_2\mX_{t-2}.
\end{aligned}
\label{eq:split-proof}
\end{equation}
Applying the source bias once recovers the original response.
If the spatial and temporal-correction matrices are approximated
by $\widetilde{\mS}$ and
$\widetilde{\mM}_{\mathrm{temp}}$, respectively,
while inputs and bias remain unchanged, the output error is
\begin{equation}
  \widetilde{\mathbf E}_t
  =(\widetilde{\mS}-\mS)\mX_t
  +(\widetilde{\mM}_{\mathrm{temp}}-\mM_{\mathrm{temp}})
    \Delta\mX_t.
  \label{eq:branch-error-decomposition}
\end{equation}
The triangle inequality and
$\|\mathbf A\mathbf B\|_{\mathrm F}
\leq\|\mathbf A\|_2\|\mathbf B\|_{\mathrm F}$
give
\begin{equation}
\begin{aligned}
  \|\widetilde{\mathbf E}_t\|_{\mathrm F}
  \leq{}&
  \|\widetilde{\mS}-\mS\|_2\|\mX_t\|_{\mathrm F}\\
  &+\|\widetilde{\mM}_{\mathrm{temp}}
    -\mM_{\mathrm{temp}}\|_2
    \|\Delta\mX_t\|_{\mathrm F}.
\end{aligned}
\label{eq:branch-error-bound}
\end{equation}
Preserving $\mS$ eliminates the spatial approximation term, leaving only the temporal-correction error.

\paragraph{Fixed-rank weight approximation.}
Let $q=\operatorname{rank}(\mM_{\mathrm{temp}})$ and write its SVD as $\mM_{\mathrm{temp}}=\sum_{i=1}^{q}\sigma_i\mathbf u_i\mathbf v_i^\top$, where $\sigma_1\geq\cdots\geq\sigma_q>0$ and $\sigma_{q+1}=0$. For $0\leq r\leq q$, truncated SVD minimizes the rank-$r$ weight-approximation error in the Frobenius norm:
\begin{equation}
  \|\mM_{\mathrm{temp},r}-\mM_{\mathrm{temp}}\|_{\mathrm F}^2
  =\sum_{i=r+1}^{q}\sigma_i^2.
  \label{eq:tlr-weight-error}
\end{equation}
The corresponding output error is input-dependent, as quantified below.

\paragraph{Local output error.}
Because $\mS$ is preserved exactly, the conversion error
at SVD initialization is
\begin{equation}
  \mathbf E_t=\widehat{\mY}_t-\mY_t
  =(\mM_{\mathrm{temp},r}
    -\mM_{\mathrm{temp}})\Delta\mX_t.
  \label{eq:tlr-error-identity}
\end{equation}
The SVD residual has spectral norm $\sigma_{r+1}$,
so submultiplicativity gives
\begin{equation}
  \|\mathbf E_t\|_{\mathrm F}
  \leq\sigma_{r+1}\|\Delta\mX_t\|_{\mathrm F}.
  \label{eq:tlr-error}
\end{equation}
For a fixed truncation, the bound scales with the historical--current feature differences. More precisely,
\begin{equation}
  \mathbf E_t
  =-\sum_{i=r+1}^{q}\sigma_i\mathbf u_i
    (\mathbf v_i^\top\Delta\mX_t).
  \label{eq:tlr-error-expansion}
\end{equation}
Orthogonality of the left singular vectors gives
\begin{equation}
  \|\mathbf E_t\|_{\mathrm F}^2
  =\sum_{i=r+1}^{q}
    \sigma_i^2
    \|\mathbf v_i^\top\Delta\mX_t\|_2^2.
  \label{eq:tlr-directional-error}
\end{equation}
Thus, the actual output error depends on both the discarded singular values and the feature-difference directions.

\paragraph{Rank monotonicity.}
For fixed source weights, input patches, and ordered SVD components,
any $0\leq r_1\leq r_2\leq q$ satisfies
\begin{equation}
  \|\mathbf E_t^{(r_1)}\|_{\mathrm F}^2
  -\|\mathbf E_t^{(r_2)}\|_{\mathrm F}^2
  =\sum_{i=r_1+1}^{r_2}
    \sigma_i^2
    \|\mathbf v_i^\top\Delta\mX_t\|_2^2
  \geq0.
  \label{eq:tlr-rank-monotonicity}
\end{equation}
Taking the spatial-only reference
$\widehat{\mY}_t^{(0)}
=\mS\mX_t+\vb\vone_P^\top$ gives
\begin{equation}
  \|\widehat{\mY}_t^{(r)}-\mY_t\|_{\mathrm F}^2
  \leq
  \|\widehat{\mY}_t^{(0)}-\mY_t\|_{\mathrm F}^2.
  \label{eq:tlr-vs-spatial-only}
\end{equation}
Each added singular component can only reduce the local conversion error; rank $q$ recovers the source layer exactly. At $r=0$, the operator reduces to the summed-slice Conv2D initialized with $\mS$.

\subsection{Rank Selection and Input Dependence}
\label{app:tlr-rank-selection}

\paragraph{Singular-value retention.}
For layers using spectral rank selection, a nonzero temporal-correction matrix and shared threshold $\eta\in(0,1]$ determine the smallest rank satisfying
\begin{equation}
  r=\min\left\{k\in\{1,\ldots,q\}:
    \frac{\sum_{i=1}^{k}\sigma_i}
         {\sum_{i=1}^{q}\sigma_i}
    \geq\eta\right\}.
  \label{eq:tlr-rank-selection}
\end{equation}
The rule retains a fraction $\eta$ of the singular-value sum and yields a layer-specific rank from each spectrum. Unless otherwise specified, $\eta=0.6$; ranks are fixed after conversion, and a zero temporal-correction matrix uses no temporal factors. The matched-MAC ablation instead allocates ranks by retained squared singular-value energy per added MAC under a prescribed decoder budget (Appendix~\ref{app:matched-mac}).

\paragraph{Dependence on video features.}
Let $\mathbf R_{\mathrm{temp}}=\mM_{\mathrm{temp},r}-\mM_{\mathrm{temp}}$ and $\mathbf C_{\Delta}=\mathbb E[\Delta\mX_t\Delta\mX_t^\top]$, where the expectation is over input videos and temporal positions. Then
\begin{equation}
  \mathbb E\|\mathbf E_t\|_{\mathrm F}^2
  =\operatorname{tr}\!\left(
    \mathbf R_{\mathrm{temp}}\mathbf C_{\Delta}
    \mathbf R_{\mathrm{temp}}^\top
  \right).
  \label{eq:tlr-expected-error}
\end{equation}
Equation~(\ref{eq:tlr-expected-error}) makes the dependence on the feature-difference covariance explicit. Figures~\ref{fig:wan21-stage-fidelity} and~\ref{fig:matched-mac} evaluate the spectral and matched-MAC protocols, respectively; Figure~\ref{fig:rank-ablation} motivates the default $\eta=0.6$.

\section{Efficient TLRConv2D: Cost and Execution}
\label{app:efficient-tlr}

\subsection{Computational Cost and Rank Condition}
\label{app:tlr-cost}

\paragraph{Counting convention.}
One multiply--accumulate (MAC) denotes an operation of the form
$a\leftarrow a+wx$.
Counts are reported for one output feature frame, with
$d=C_{\mathrm{in}}k_hk_w$ and
$P=H_{\mathrm{out}}W_{\mathrm{out}}$.
Standalone additions, feature differences, bias additions, nonlinearities, and memory transfers are excluded. Counting multiplication and addition separately gives approximately two floating-point operations per MAC. For $N$ videos with $T_{\mathrm{out}}$ output frames, the counts scale by $NT_{\mathrm{out}}$; network-level totals sum the actual feature shapes and ranks of all layers. The subscript $\mathrm{TLR}$ identifies the computational cost of the TLRConv2D operator.

\paragraph{Source and low-rank operators.}
Each temporal kernel slice costs
$PC_{\mathrm{out}}d$ MACs, so
\begin{equation}
  c_{\mathrm{3D}}
  =\sum_{j=0}^{2}PC_{\mathrm{out}}d
  =3PC_{\mathrm{out}}d.
  \label{eq:source-mac-derivation}
\end{equation}
The exact spatial-response/temporal-correction rewriting alone does not change this count. TLRConv2D retains the dense spatial response, computes two rank-$r$ temporal projections, and applies one shared pointwise output projection:
\begin{equation}
\begin{aligned}
  c_{\mathrm{TLR}}={}&
    \underbrace{
      PC_{\mathrm{out}}d
    }_{\text{spatial response}}
    +\underbrace{
      2Prd
    }_{\text{temporal projections}}\\
    &+\underbrace{
      PC_{\mathrm{out}}r
    }_{\text{pointwise projection}}.
\end{aligned}
\label{eq:tlr-mac-derivation}
\end{equation}
The two projected temporal terms are summed before applying $\mA$; applying $\mA$ to them separately would require two pointwise projections. For $P>0$,
\begin{equation}
\begin{aligned}
  c_{\mathrm{TLR}}<c_{\mathrm{3D}}
  &\iff
  C_{\mathrm{out}}d+2rd+C_{\mathrm{out}}r
  <3C_{\mathrm{out}}d\\
  &\iff
  r(2d+C_{\mathrm{out}})
  <2C_{\mathrm{out}}d.
\end{aligned}
\label{eq:tlr-rank-condition-derivation}
\end{equation}
Thus, TLRConv2D uses fewer convolutional MACs when
\begin{equation}
  r<
  \frac{2C_{\mathrm{out}}d}
       {2d+C_{\mathrm{out}}},
  \label{eq:tlr-rank-condition}
\end{equation}
and its arithmetic ratio is
\begin{equation}
  \frac{c_{\mathrm{TLR}}}{c_{\mathrm{3D}}}
  =\frac{1}{3}
   +\frac{2r}{3C_{\mathrm{out}}}
   +\frac{r}{3d}.
  \label{eq:tlr-mac-ratio}
\end{equation}
The spatial cost remains fixed, while the temporal cost grows with rank. These expressions quantify convolutional arithmetic; measured GPU latency also depends on execution overhead and memory traffic.

\subsection{Folding Equivalence and History Indexing}
\label{app:tlr-folding}

\paragraph{Collecting current-frame terms.}
Expanding Equation~(\ref{eq:tlr}) gives
\begin{equation}
\begin{aligned}
  \widehat{\mY}_t={}&
    \mS\mX_t+\vb\vone_P^\top
    +\mA\mB_1\mX_{t-1}
    +\mA\mB_2\mX_{t-2}\\
    &-\mA\mB_1\mX_t
    -\mA\mB_2\mX_t.
\end{aligned}
\label{eq:tlr-expanded-forward}
\end{equation}
Collecting the terms involving $\mX_t$ yields
\begin{equation}
\begin{aligned}
  \widehat{\mY}_t={}&
    [\mS-\mA(\mB_1+\mB_2)]\mX_t
    +\vb\vone_P^\top\\
    &+\mA(
      \mB_1\mX_{t-1}
      +\mB_2\mX_{t-2}
    ).
\end{aligned}
\label{eq:folded-tlr-derivation}
\end{equation}
The bracketed term is $\mS_{\mathrm{eff}}$. Because $\mA$ is pointwise and all factors are linear, each product $\mA\mB_j$ contracts only the intermediate channel dimension and does not enlarge the spatial kernel. Folding is performed after recovery and introduces no additional approximation or rank truncation.

\paragraph{Shared-input joint convolution.}
Equation~(\ref{eq:tlr-joint-convolution}) stacks
$\mS_{\mathrm{eff}}$, $\mB_1$, and $\mB_2$
along the output-channel dimension.
The kernels share the same spatial stride, dilation, and padding; their output widths are $C_{\mathrm{out}}$, $r$, and $r$, respectively, and only the first output group receives the source bias. The joint Conv2D and pointwise history projection cost
$(C_{\mathrm{out}}+2r)dP$ and
$C_{\mathrm{out}}rP$ MACs, respectively,
whose sum equals $c_{\mathrm{TLR}}$.

\paragraph{History indexing.}
For an incoming frame at time $s$,
$\mathbf Q_{j,s}=\mB_j\mX_s$
records both the projection kernel $j$ and
the source time $s$. Therefore,
\begin{equation}
  \mB_1\mX_{t-1}=\mathbf Q_{1,t-1},
  \qquad
  \mB_2\mX_{t-2}=\mathbf Q_{2,t-2}.
  \label{eq:tlr-lagged-projections}
\end{equation}
Away from the sequence boundary, the current projections are used only by subsequent outputs because their negative current-frame contribution is already folded into $\mS_{\mathrm{eff}}$:
\begin{equation}
\begin{aligned}
  \widehat{\mY}_{t+1}
  &=\mathbf R_{t+1}
    +\mA(
      \mathbf Q_{1,t}
      +\mathbf Q_{2,t-1}
    ),\\
  \widehat{\mY}_{t+2}
  &=\mathbf R_{t+2}
    +\mA(
      \mathbf Q_{1,t+1}
      +\mathbf Q_{2,t}
    ).
\end{aligned}
\label{eq:tlr-projection-future-use}
\end{equation}
Thus, $\mathbf Q_{1,t}$ is required one step later and $\mathbf Q_{2,t}$ two steps later; $\mathbf Q_{2,t-1}$ also remains available for the next output.

At each step, execution proceeds as follows:
\begin{enumerate}
  \item Compute $\mathbf R_t$, $\mathbf Q_{1,t}$,
  and $\mathbf Q_{2,t}$ using the joint Conv2D.

  \item Read $\mathbf Q_{1,t-1}$ and
  $\mathbf Q_{2,t-2}$ according to the source
  temporal-padding rule.

  \item Apply $\mA$ to their sum and add the result
  to $\mathbf R_t$.

  \item Update the cache after all entries required
  by the current output have been read.
\end{enumerate}
The cache persists across chunks of the same video and is reset between independent videos. At sequence start, zero padding creates zero virtual entries. Under replicated-first-frame padding, projections of the first input frame populate the virtual history before the first output is formed.

\paragraph{Execution efficiency.}
Folding replaces three shared-input spatial-convolution calls with one joint call while retaining the pointwise history projection and the TLR MAC count. With matching bias placement and temporal boundary handling, it is algebraically equivalent up to floating-point evaluation order. Figure~\ref{fig:operator-ablation} reports the resulting throughput and peak memory.

\section{Block-Pruning Implementation}
\label{app:search}

We adapt TinyFusion's recoverability-aware search~\citep{fang2024tinyfusion} to video autoencoders, using grouped block selection, shared LoRA adaptation, and straight-through Gumbel-Softmax estimation.

\paragraph{Search objective.}
For decoder search on a video $\tV$, the teacher encoder supplies posterior-mean latents $\tZ=\mathcal E_{\mathrm T}(\tV)$, and a sampled structure produces $\tY_{\mathrm{search}}$. Selection logits and shared LoRA parameters are optimized with frozen base weights using
\begin{equation}
\begin{aligned}
  \mathcal L_{\mathrm{search}}
  =\mathbb E_{\tV,\boldsymbol\xi}\bigl[
    &\mathcal L_1(\tY_{\mathrm{search}},\tV)\\
    &+\lambda_{\mathrm{perc}}^{\mathrm{search}}
      \operatorname{LPIPS}(\tY_{\mathrm{search}},\tV)
  \bigr],
\end{aligned}
\label{eq:search-loss}
\end{equation}
where $\boldsymbol\xi$ denotes structure-sampling noise and $\lambda_{\mathrm{perc}}^{\mathrm{search}}$ weights the search-time perceptual loss.

\subsection{Selection, Adaptation, and Export}
\label{app:search-implementation}

\paragraph{Legal pruning units.}
For a convolutional residual block, a binary retention
gate $m_\ell$ gives
\begin{equation}
  \tH_{\ell+1}
  =\mathcal P_\ell(\tH_\ell)
   +m_\ell\mathcal F_\ell(\tH_\ell),
  \label{eq:residual-gating}
\end{equation}
where $\mathcal P_\ell$ includes any required shortcut projection. Setting $m_\ell=0$ removes the residual computation while retaining the shortcut path. Compatible Transformer blocks may be replaced by the identity. Resampling layers and network input/output layers remain fixed, preserving feature dimensions at retained-block interfaces.

\paragraph{Grouped selection.}
For a group of $n_g$ blocks with retention quota $k_g$, the legal masks are $\mathcal M_g=\{\mathbf m\in\{0,1\}^{n_g}:\sum_jm_j=k_g\}$. Each candidate mask has a learned selection logit. Straight-through Gumbel-Softmax~\citep{jang2016gumbel} with temperature $\tau>0$ samples one legal mask in the forward pass and uses relaxed probabilities for the backward pass. All groups are optimized jointly using Equation~(\ref{eq:search-loss}).

\paragraph{Adaptation and export.}
The pretrained base weights remain frozen while selection logits and shared LoRA parameters~\citep{hu2021lora} are optimized. After search, each group retains the highest-scoring mask according to the exponential moving average of its logits. LoRA updates are merged into retained weights as
\begin{equation}
  \mW_{\mathrm{merge}}
  =\mW+
  \frac{\alpha_{\mathrm L}}{r_{\mathrm L}}\mL\mR,
  \label{eq:lora-merge}
\end{equation}
where $\mL$ and $\mR$ are the learned LoRA factors, with shapes $m\times r_{\mathrm L}$ and $r_{\mathrm L}\times n$ for a reshaped weight $\mW\in\mathbb R^{m\times n}$. The symbols $r_{\mathrm L}$ and $\alpha_{\mathrm L}$ denote the LoRA rank and scaling coefficient, respectively. Eligible merged weights are then converted using TLR, followed by probe-guided reconstruction recovery and folded export. Block selection preserves the widths of retained representations, while TLR controls temporal capacity through layer-specific ranks. Depth selection and temporal-rank allocation are performed sequentially.

\paragraph{Applicability.}
\label{sec:lightvae-scope}
Block search applies to compatible convolutional and Transformer blocks in both encoders and decoders. In our experiments, TLR is applied to the Wan convolutional VAEs, whereas MiniMax-H3 uses block pruning only. Decoder reconstruction uses teacher-encoder latents; encoder compression is evaluated in downstream reference-conditioned generation.

\subsection{Feature Visualization}
\label{app:pruning-feature-visualization}

\begin{figure}[t]
  \centering
  \includegraphics[width=0.96\linewidth]{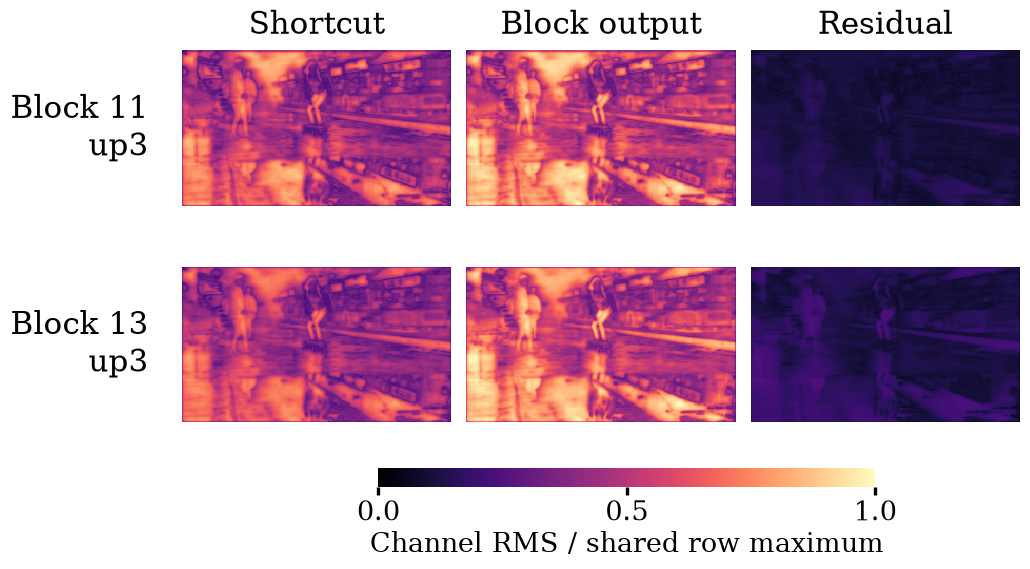}
  \caption{\textbf{Single-video feature example.}
  Channel-wise RMS of the shortcut, block output,
  and residual for blocks 11 and 13 at feature-frame
  index 8; each row uses a shared scale.
  Blocks are selected using aggregate search scores
  rather than this video's features.}
  \label{fig:pruning-features}
\end{figure}

Figure~\ref{fig:pruning-features} visualizes the shortcut output, full block output, and residual response for two high-similarity blocks from one video. The maps show channel-wise RMS at feature-frame index 8, with a shared scale within each row. These examples illustrate the small residual response of high-similarity blocks. Pruning decisions use aggregate search scores over the calibration data.

\section{Teacher-Suffix Probe and Feature Alignment}
\label{app:probe}

During recovery, we cycle through compatible group boundaries and probe one boundary per iteration.

\paragraph{Teacher-suffix probe.}
For the same latent $\tZ$, let
$\tY_q=\mathcal D_q(\tZ)$
for $q\in\{\mathrm T,\mathrm S\}$.
At a compatible intermediate boundary, write
$\mathcal D_q=
\mathcal D_q^{>}\circ\mathcal D_q^{\leq}$
and
$\tH_q=\mathcal D_q^{\leq}(\tZ)$.
The teacher and student features have compatible shapes at this boundary.
The frozen teacher suffix probes the student feature as
$\tY_{\mathrm P}
=\mathcal D_{\mathrm T}^{>}(\tH_{\mathrm S})$.
With
$\mathbf h_q=\operatorname{vec}(\tH_q)
\in\mathbb R^{N_f}$, define
\begin{equation}
\begin{aligned}
  \mathcal L_{\mathrm P}
  &=\frac{1}{N_y}
    \left\|\operatorname{vec}
    \bigl(
      \tY_{\mathrm P}
      -\operatorname{sg}(\tY_{\mathrm T})
    \bigr)
    \right\|_2^2,\\
  \mathbf g
  &=\nabla_{\mathbf h_{\mathrm S}}\mathcal L_{\mathrm P},
  \qquad
  \boldsymbol\delta
  =\mathbf h_{\mathrm S}
   -\operatorname{sg}(\mathbf h_{\mathrm T}).
\end{aligned}
\label{eq:teacher-suffix-probe}
\end{equation}
Here, $N_f$ and $N_y$ are the numbers of feature and output elements, respectively. The suffix parameters are frozen while gradients remain enabled at its input.

\paragraph{Probe-guided feature alignment.}
Ordinary feature matching moves coordinate $i$ along $-\delta_i$. When $\delta_i g_i<0$, this coordinate-wise direction locally increases the probe loss. We suppress such directions and weight the remaining feature errors by
\begin{equation}
\begin{aligned}
  w_i
  &=\operatorname{clip}_{[0,w_{\max}]}
    \left(
      \frac{[\delta_i(N_f g_i/2)]_+}
           {\delta_i^2+\rho}
    \right),\\
  \mathcal L_{\mathrm{feature}}
  &=\frac{1}{N_f}\sum_{i=1}^{N_f}
    \operatorname{sg}(w_i)\delta_i^2.
\end{aligned}
\label{eq:probe-weighted-feature}
\end{equation}
Here, $w_i$ is the nonnegative alignment weight for coordinate $i$, $[a]_+=\max(a,0)$, $\rho>0$ stabilizes the denominator, and $w_{\max}>0$ caps the weights. The factor $N_f/2$ accounts for the normalization of the feature-MSE gradient. The weights are detached during student optimization, so no second-order differentiation is required.

\paragraph{Comparison with ordinary feature MSE.}
The unweighted feature-matching baseline is
\begin{equation}
  \mathcal L_{\mathrm{feature}}^{\mathrm{MSE}}
  =\frac{1}{N_f}\sum_{i=1}^{N_f}\delta_i^2.
  \label{eq:feature-mse-baseline}
\end{equation}
Using the reconstruction objective
$\mathcal L_{\mathrm{recon}}$
defined in Equation~(\ref{eq:recovery-loss}),
the three recovery objectives are
\begin{equation}
\begin{aligned}
  &\mathcal L_{\mathrm{recon}},\\
  &\mathcal L_{\mathrm{recon}}
   +\lambda_{\mathrm F}
    \mathcal L_{\mathrm{feature}}^{\mathrm{MSE}},\\
  &\mathcal L_{\mathrm{restore}}
   =\mathcal L_{\mathrm{recon}}
   +\lambda_{\mathrm F}
    \mathcal L_{\mathrm{feature}}.
\end{aligned}
\label{eq:recovery-ablation-objectives}
\end{equation}
The second objective assigns equal weights to feature coordinates; the third uses detached probe-guided weights. All three objectives use the same architecture with 5 of 14 residual main branches retained, initialization, data, recovery budget, and feature boundaries. Both feature objectives use the fixed coefficient $\lambda_{\mathrm F}=0.01$. Under this controlled comparison, Figure~\ref{fig:recovery-ablation} shows that uniform feature MSE underperforms reconstruction alone, whereas probe-guided alignment improves recovery. The probe loss is used only to construct detached weights.

\paragraph{Feature-space alignment.}
At a selected boundary, let
$\mathbf u=
\nabla_{\mathbf h_{\mathrm S}}
\mathcal L_{\mathrm{feature}}$.
Because the weights are detached during student optimization,
\begin{equation}
  u_i=\frac{2}{N_f}w_i\delta_i.
  \label{eq:probe-feature-gradient}
\end{equation}
If $\delta_i g_i\leq0$, then $w_i=0$.
Otherwise, $u_i$ and $g_i$ have the same sign.
Consequently,
\begin{equation}
  \mathbf g^\top\mathbf u
  =\frac{2}{N_f}
    \sum_iw_i\delta_i g_i
  \geq0.
  \label{eq:probe-alignment}
\end{equation}
Thus, the feature-loss gradient is nonnegatively aligned with the probe-loss gradient in feature space.

\paragraph{Damping and clipping.}
For a coordinate satisfying $\delta_i g_i>0$,
before upper clipping,
\begin{equation}
  u_i
  =g_i\frac{\delta_i^2}{\delta_i^2+\rho}.
  \label{eq:probe-damped-gradient}
\end{equation}
Damping attenuates updates for small discrepancies, and clipping limits their magnitude without changing the sign relation. In the zero-damping, unclipped limit, aligned coordinates pass the probe gradient and all other coordinates receive zero weight.

\paragraph{Local feature-space alignment.}
For the hypothetical feature-space update
$\mathbf h_{\mathrm S}'
=\mathbf h_{\mathrm S}-\epsilon\mathbf u$,
differentiability gives
\begin{equation}
  \mathcal L_{\mathrm P}(\mathbf h_{\mathrm S}')
  =\mathcal L_{\mathrm P}(\mathbf h_{\mathrm S})
   -\epsilon\mathbf g^\top\mathbf u
   +o(\epsilon).
  \label{eq:probe-local-step}
\end{equation}
Equation~(\ref{eq:probe-alignment}) therefore guarantees a nonpositive first-order change when $\mathbf h_{\mathrm S}$ is updated directly. For a parameter update with student-prefix Jacobian $\mathbf J=\partial\mathbf h_{\mathrm S}/\partial\boldsymbol\theta$, the gradient interaction becomes
\begin{equation}
  (\mathbf J^\top\mathbf g)^\top
  (\mathbf J^\top\mathbf u)
  =
  \mathbf g^\top
  \mathbf J\mathbf J^\top
  \mathbf u.
  \label{eq:probe-parameter-alignment}
\end{equation}
Because $\mathbf J\mathbf J^\top$ couples feature coordinates, the feature-space guarantee need not hold in parameter space; Figure~\ref{fig:recovery-ablation} evaluates the resulting training effect.

\section{Experimental Details}
\label{app:training-settings}

\subsection{Training and Recovery}
\label{app:optimization-settings}

\paragraph{Search and optimization.}
Block search follows TinyFusion~\citep{fang2024tinyfusion}, using LoRA~\citep{hu2021lora} with $r_{\mathrm L}=16$ and $\alpha_{\mathrm L}=32$. Search and recovery share AdamW~\citep{loshchilov2019decoupled} with $(\beta_1,\beta_2)=(0.9,0.999)$, zero weight decay, and a per-GPU batch size of one without gradient accumulation. The learning rate warms up to $10^{-5}$ over 50 steps and then follows cosine decay to $10^{-6}$. We clip the gradient norm to 1.0 and use an exponential moving average decay of 0.999.

\paragraph{Recovery loss coefficients.}
In Equation~(\ref{eq:recovery-loss}), we use fixed coefficients $\lambda_{\mathrm{perc}}=0.8$, $\lambda_{\mathrm{adv}}=0.05$, and $\lambda_{\mathrm F}=0.01$ throughout recovery. Uniform feature MSE and probe-guided alignment use the same feature coefficient and feature boundaries. The coordinate weights $w_i$ are computed from probe gradients and signed feature discrepancies, as defined in Appendix~\ref{app:probe}.

\subsection{Compression and Ablation Settings}
\label{app:compression-settings}

\paragraph{Model configurations.}
For the Wan VAEs, LightVAE-Pro retains all 14 residual blocks, while LightVAE-Lite retains one residual main branch per stage ($14\!\rightarrow\!5$). Both use $r=0$ in the converted residual convolutions of \texttt{up2}/\texttt{up3}; elsewhere, TLR selects layer-specific ranks using $\eta=0.6$. MiniMax-H3 uses block pruning only: $36\!\rightarrow\!12$ decoder Transformer blocks and $12\!\rightarrow\!6$ encoder residual main branches. Required shortcuts, projections, and resampling operations are retained.

\paragraph{Operator comparisons.}
Operator ablations preserve the original Wan2.1 decoder depth and width. Conv2D uses summed temporal kernel slices. The DW3D+PW/Conv2D hybrids follow Flash-VAED's operator layout~\citep{zhu2026flashvaed}, with random-DW or teacher-SVD initialization, excluding its channel pruning and training recipe. For teacher-SVD initialization, we apply a rank-one SVD to each input channel's pretrained Conv3D kernel, extending the DW+PW decomposition of \citet{guo2018nd} to three dimensions. All compressed variants use the same recovery protocol; Figure~\ref{fig:operator-ablation} reports reconstruction quality, throughput, and peak memory.

\paragraph{Spatial--temporal approximation protocols.}
\label{app:matched-mac}
Figures~\ref{fig:wan21-stage-fidelity} and~\ref{fig:matched-mac} use 1,024 and 256 Wan2.1-generated videos, respectively, retaining the original decoder depth and width without recovery. For each converted convolution, $\mS=\mW_0+\mW_1+\mW_2$ acts on current features, while $\mM_{\mathrm{temp}}=[\mW_1,\mW_2]$ acts on historical--current feature differences. The stage-wise comparison modifies one stage's residual convolutions at a time using the same singular-value retention threshold $\eta=0.6$; other stages remain unchanged. The matched-MAC comparison uses 81-frame $480\times832$ clips and includes the same 28 residual convolutions in the Wan2.1 decoder. All candidates use identical original-encoder latents. We report the equal-video mean of $\|\hat y_i-y_i^{\mathrm{orig}}\|_2/\|y_i^{\mathrm{orig}}\|_2$, measuring final-output deviation from the original decoder before clamping.

\paragraph{Budget-dependent rank allocation.}
For matched-MAC comparisons, each compressed branch starts at rank one. Additional ranks are allocated greedily by the gain in retained squared singular-value energy per added MAC, independently for each budget using source weights only. Spatial-only compression retains dense $\mM_{\mathrm{temp}}$, while TLR retains dense $\mS$; both configurations allow individual compressed branches to return to dense execution. For joint compression, we divide the required absolute MAC savings equally between the spatial and temporal branches and allocate ranks within each branch budget independently.

\paragraph{MAC limits and curve interpretation.}
The original decoder costs approximately 137.68 TMAC per video. Fixed operations, dense $\mS$, and dense $\mM_{\mathrm{temp}}$ account for 8.86\%, 30.38\%, and 60.76\%, respectively. Counting convolutions and attention matrix products, the rank-one minimum yields MAC floors of 69.86\% for spatial-only compression, 39.70\% for TLR, and 39.72\% for the equal-saving joint baseline. Consequently, spatial-only compression is evaluated only at 70\%, 80\%, 90\%, and 95\%; TLR and joint compression additionally cover 40\%, 50\%, and 60\%. The plot uses actual MAC ratios. At a 70\% total budget, spatial-only and joint compression retain only 1.26\% and 50.63\% of the original spatial MACs, respectively, consistent with their errors of 1.1585 and 0.6965. At 80\%, spatial-only compression restores its spatial budget to 34.17\% while retaining the full temporal branch, and its error drops below joint compression. Under this allocation, TLR yields the lowest error at every evaluated budget. This advantage is consistent with preserving the dense current-frame spatial branch $\mS$ while concentrating approximation on the temporal correction.

\paragraph{Joint compression.}
Table~\ref{tab:compression-complementarity} evaluates block pruning and TLR separately and jointly on NVIDIA H100 and RTX~5090 using the Wan input settings of Table~\ref{tab:main-results}. Pruning retains 5 of 14 residual main branches. In TLR configurations without the zero-rank setting, we use $\eta=0.6$ in all stages. The zero-rank setting sets $r=0$ in \texttt{up2}/\texttt{up3} and uses $\eta=0.6$ elsewhere. The first and last rows of Table~\ref{tab:compression-complementarity} therefore match the original and LightVAE-Lite configurations in Table~\ref{tab:main-results}, respectively.

\paragraph{Depth and channel pruning.}
For the pruning comparison, both variants retain dense Conv3D: block pruning keeps 6 of 14 residual main branches (57.1\% removal), while channel pruning removes approximately 41.7\% of channels. Their pre-recovery RTX~5090 throughput is 45.58 and 42.69 FPS, respectively, enabling comparison at similar inference cost under the same recovery protocol. The separate feature-loss ablation fixes the architecture at 5 of 14 retained residual main branches (Appendix~\ref{app:probe}).

\subsection{Inference Benchmarks}
\label{app:inference-benchmarks}

The decoding benchmarks use NVIDIA H100 and GeForce RTX~5090 GPUs with Python 3.12.3, PyTorch \texttt{2.11.0+cu130}, CUDA 13.0, and cuDNN 9.19.0. Table~\ref{tab:main-results} uses 81-frame clips at $480\times832$ for Wan2.1 and $704\times1280$ for Wan2.2, and 124-frame $768\times1344$ clips for MiniMax-H3. The joint-compression ablation in Table~\ref{tab:compression-complementarity} uses both GPUs and the same Wan input settings as Table~\ref{tab:main-results}. These decoding-throughput settings differ from VBench-2.0 generation in Section~\ref{sec:experimental-setup}, where Wan2.2 uses 121 frames. OOM denotes an out-of-memory failure and is not assigned a numerical throughput or speedup.


\section{Additional Downstream Evaluation}
\label{app:downstream-evaluation}

\subsection{Fine-Grained VBench-2.0 Results}
\label{app:vbench2-breakdown}

Table~\ref{tab:vbench2-fine} reports all 18 VBench-2.0 dimensions for Wan2.1, Wan2.2, and MiniMax-H3 on a 0--100 scale. Videos follow the matched generation protocol in Section~\ref{sec:experimental-setup}: only the decoder varies within each backbone.
On Wan2.1, Pro scores 53.1251 versus 52.9867 for the original decoder, while Lite retains 98.52\% of the original overall score. On Wan2.2, Pro and Lite score 57.5155 and 56.9366, respectively, compared with 56.8765 for the original decoder and 57.7332 for Turbo-VAED. On MiniMax-H3, Lite retains 99.51\% of the original overall score. The evaluated efficient decoders maintain aggregate generation scores close to the originals despite their reconstruction differences, suggesting limited sensitivity of the overall score under these compression settings. Individual dimensions can still change substantially: H3 Lite improves Motion Rationality but decreases Multi-View Consistency.

\begin{table*}[t]
\centering
\caption{Fine-grained VBench-2.0 generation quality (0--100; higher is better). Metrics continue in the lower panel. Dashed lines separate original and compressed decoders. Official Overall uses the benchmark aggregation.}
\label{tab:vbench2-fine}
\begingroup
\small
\setlength{\tabcolsep}{3pt}
\setlength{\dashlinedash}{2pt}
\setlength{\dashlinegap}{2pt}
\renewcommand{\arraystretch}{1.2}
\resizebox{\textwidth}{!}{%
\begin{tabular}{@{}ll*{10}{c}@{}}
\toprule
\shortstack[l]{Generation\\backbone} & Decoder
& \shortstack{Camera\\Motion} & \shortstack{Complex\\Landscape} & \shortstack{Complex\\Plot} & \shortstack{Composition} & \shortstack{Diversity} & \shortstack{Dynamic\\Attribute} & \shortstack{Dynamic\\Spatial\\Relationship} & \shortstack{Human\\Anatomy} & \shortstack{Human\\Clothes} & \shortstack{Human\\Identity} \\
\midrule
\multirow{4}{*}{Wan2.1-1.3B} & Original & 16.0494 & 16.8889 & 10.4722 & 41.9706 & 62.6306 & 13.5531 & 21.2560 & 88.7595 & 94.5701 & 72.6868 \\
\cdashline{2-12}
 & Flash-VAED & 14.1975 & 17.7778 & 9.5556 & 42.3375 & 56.9775 & 12.8205 & 21.7391 & 86.2864 & 93.6652 & 69.4928 \\
 & LightVAE-Pro & 16.0494 & 18.6667 & 8.7778 & 42.0755 & 61.7671 & 13.5531 & 23.1884 & 87.8389 & 94.5701 & 72.1133 \\
 & LightVAE-Lite & 15.4321 & 19.1111 & 9.0556 & 41.1321 & 55.6720 & 13.9194 & 20.7729 & 87.0935 & 94.0909 & 71.3987 \\
\midrule
\multirow{4}{*}{Wan2.2-5B} & Original & 10.8025 & 20.4444 & 11.3611 & 46.0901 & 67.1367 & 10.9890 & 25.1208 & 87.4368 & 97.6744 & 73.9860 \\
\cdashline{2-12}
 & Turbo-VAED & 11.1111 & 21.1111 & 12.9167 & 46.2683 & 70.0579 & 12.0879 & 25.1208 & 87.2452 & 98.1481 & 73.6227 \\
 & LightVAE-Pro & 10.8025 & 20.4444 & 12.2500 & 47.6520 & 67.5512 & 11.3553 & 25.1208 & 87.0102 & 98.1308 & 74.1267 \\
 & LightVAE-Lite & 10.8025 & 22.0000 & 11.9167 & 47.0231 & 64.4702 & 10.9890 & 25.6039 & 86.9995 & 98.1395 & 72.6945 \\
\midrule
\multirow{2}{*}{MiniMax-H3} & Original & 70.9877 & 26.4444 & 21.0370 & 60.8700 & 30.4177 & 82.4176 & 52.6570 & 95.4118 & 94.6429 & 61.8515 \\
\cdashline{2-12}
 & LightVAE-Lite & 72.5309 & 26.4444 & 21.6204 & 61.9497 & 28.9380 & 82.0513 & 54.1063 & 92.6465 & 93.3333 & 61.8369 \\
\bottomrule
\end{tabular}%
}

\par\medskip

\resizebox{\textwidth}{!}{%
\begin{tabular}{@{}ll*{9}{c}@{}}
\toprule
\shortstack[l]{Generation\\backbone} & Decoder
& \shortstack{Human\\Interaction} & \shortstack{Instance\\Preservation} & \shortstack{Material} & \shortstack{Mechanics} & \shortstack{Motion Order\\Understanding} & \shortstack{Motion\\Rationality} & \shortstack{Multi-View\\Consistency} & \shortstack{Thermotics} & \shortstack{Official\\Overall} \\
\midrule
\multirow{4}{*}{Wan2.1-1.3B} & Original & 54.6667 & 91.8129 & 28.9157 & 53.9683 & 19.5286 & 31.6092 & 39.7984 & 52.5547 & 52.9867 \\
\cdashline{2-11}
 & Flash-VAED & 53.6667 & 91.8129 & 30.5882 & 57.6000 & 20.5387 & 31.0345 & 46.4271 & 52.5547 & 52.4985 \\
 & LightVAE-Pro & 54.0000 & 91.8129 & 29.0698 & 54.0984 & 18.8552 & 30.4598 & 48.0858 & 52.1739 & 53.1251 \\
 & LightVAE-Lite & 53.3333 & 91.8129 & 29.4118 & 57.1429 & 21.2121 & 31.6092 & 41.7962 & 51.0949 & 52.2005 \\
\midrule
\multirow{4}{*}{Wan2.2-5B} & Original & 60.0000 & 91.8129 & 39.7436 & 59.1241 & 19.8653 & 38.5057 & 57.3342 & 58.1560 & 56.8765 \\
\cdashline{2-11}
 & Turbo-VAED & 59.3333 & 91.8129 & 38.9610 & 57.6642 & 21.2121 & 40.2299 & 63.5777 & 59.2857 & 57.7332 \\
 & LightVAE-Pro & 61.3333 & 91.8129 & 38.4615 & 59.5588 & 20.2703 & 38.5057 & 65.3694 & 57.8571 & 57.5155 \\
 & LightVAE-Lite & 60.6667 & 91.8129 & 38.4615 & 58.6957 & 20.2020 & 38.5057 & 62.0777 & 59.4203 & 56.9366 \\
\midrule
\multirow{2}{*}{MiniMax-H3} & Original & 81.6667 & 93.5673 & 89.6552 & 78.7234 & 45.7912 & 50.5747 & 63.8827 & 80.9524 & 66.8832 \\
\cdashline{2-11}
 & LightVAE-Lite & 83.0000 & 93.5673 & 89.6552 & 80.5755 & 47.1380 & 54.5977 & 51.9436 & 79.3103 & 66.5546 \\
\bottomrule
\end{tabular}%
}
\endgroup
\end{table*}

\subsection{Reference-Conditioned Generation}
\label{app:h3-ref2v}

We evaluate joint encoder--decoder compression in MiniMax-H3 reference-conditioned generation. LightVAE-Lite reduces the encoder from 12 to 6 residual main branches and the decoder from 36 to 12 Transformer blocks, while retaining required shortcuts, projections, and resampling operations. We compare the original encoder and decoder with their compressed counterparts under five reference-input configurations: one, three, six, or nine reference images, and nine images together with one reference video.

Table~\ref{tab:h3-ref2v} reports encoding and decoding latency separately. Encoding latency covers visual-VAE encoding of all reference images and the reference video when present, excluding other reference-conditioning modules. Decoding latency covers VAE decoding of the generated latents. The original and compressed pipelines use identical reference inputs, prompts, random seeds, diffusion backbones, and sampling settings. Output resolution and video duration are fixed across configurations. Speedup is computed as the original latency divided by the corresponding LightVAE-Lite latency.

Reference encoding is accelerated by $2.12\times$--$2.68\times$, and decoding by $2.68\times$--$2.75\times$ in this pipeline. Figure~\ref{fig:h3-ref-visualization} compares the jointly compressed encoder--decoder pipeline with the original pipeline across four examples. The selected frames remain visually close in overall appearance and scene content, supporting the practical extension to reference encoding.

\begin{table*}[t]
\centering
\caption{\textbf{Encoding and decoding efficiency for MiniMax-H3 reference-conditioned generation.}
Original denotes the uncompressed encoder and decoder, whereas Lite uses their
compressed counterparts. All reference images, reference-video frames, and
decoded output frames have a spatial resolution of $768\times1344$.
Both the input reference video and the decoded output video contain 124 frames.
Latency is measured in seconds on a single NVIDIA H100 GPU. Speedup is computed
as the Original latency divided by the corresponding Lite latency.}
\label{tab:h3-ref2v}
\begingroup
\small
\setlength{\tabcolsep}{7pt}
\renewcommand{\arraystretch}{1.15}
\resizebox{\textwidth}{!}{%
\begin{tabular}{@{}lcccccc@{}}
\toprule
\multirow{2}{*}{Reference input}
& \multicolumn{3}{c}{Encoding}
& \multicolumn{3}{c}{Decoding} \\
\cmidrule(lr){2-4}
\cmidrule(l){5-7}

& Original (s)$\downarrow$
& Lite (s)$\downarrow$
& Speedup$\uparrow$
& Original (s)$\downarrow$
& Lite (s)$\downarrow$
& Speedup$\uparrow$ \\
\midrule

1 image
& 1.6925 & 0.6838 & 2.48$\times$ & 8.6724 & 3.1511 & 2.75$\times$ \\

3 images
& 5.1713 & 1.9299 & 2.68$\times$ & 8.6762 & 3.1553 & 2.75$\times$ \\

6 images
& 10.3361 & 4.0597 & 2.55$\times$ & 8.6828 & 3.1932 & 2.72$\times$ \\

9 images
& 14.8763 & 5.9142 & 2.52$\times$ & 8.5608 & 3.1933 & 2.68$\times$ \\
\midrule

9 images + 1 video
& 28.8845 & 13.6354 & 2.12$\times$ & 8.6418 & 3.1761 & 2.72$\times$ \\

\bottomrule
\end{tabular}%
}
\endgroup

\par\bigskip
\includegraphics[width=\textwidth]{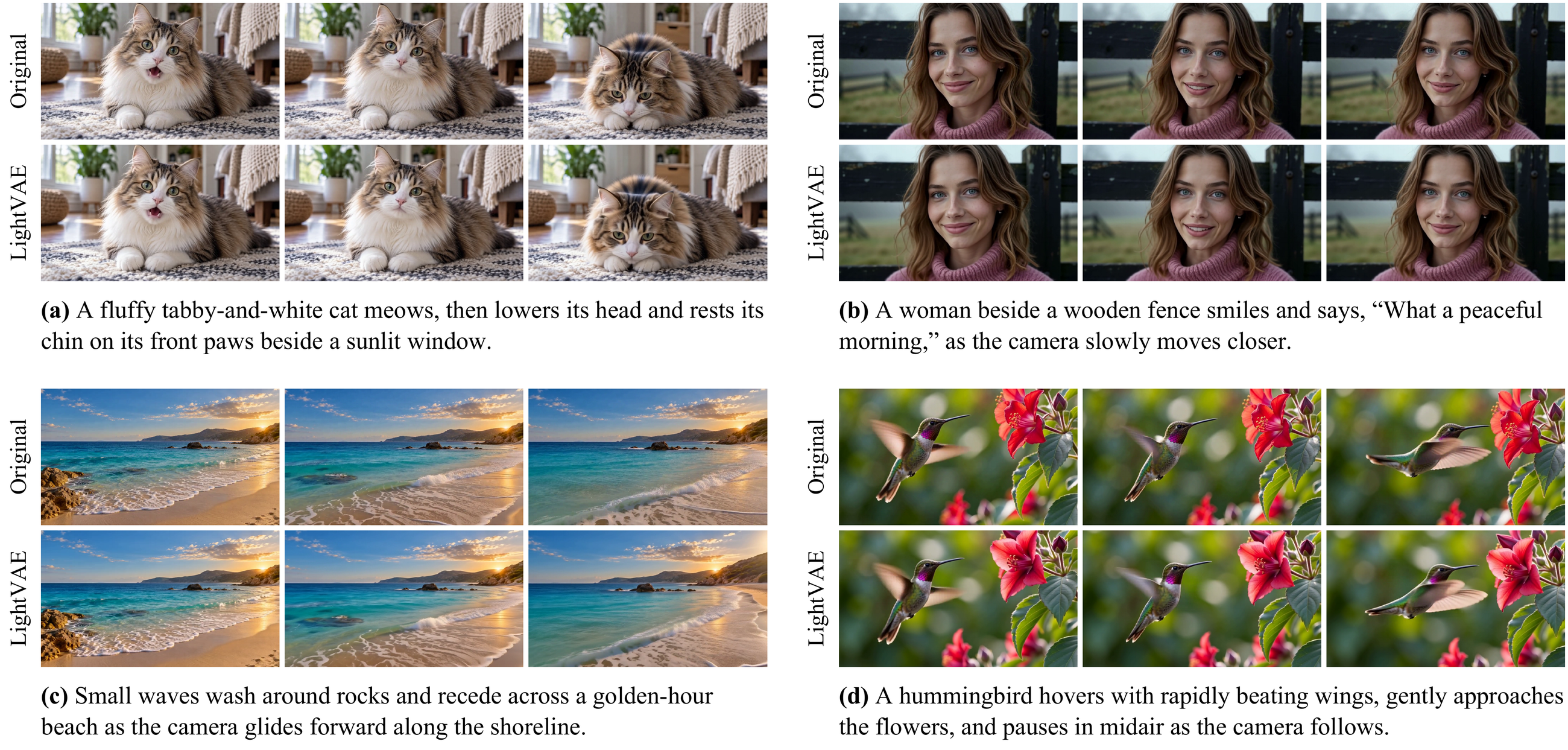}
\begingroup
\makeatletter
\def\@captype{figure}
\makeatother
\caption{\textbf{Qualitative comparison in MiniMax-H3 reference-conditioned generation.}
Original (top) uses the original encoder and decoder; LightVAE (bottom) compresses both. Each example shows three frames in temporal order with its prompt summary below. The displayed outputs remain visually close in overall appearance and scene content.}
\label{fig:h3-ref-visualization}
\endgroup
\end{table*}

\end{document}